\documentclass[twoside]{article}
\usepackage[accepted]{aistats2023}

\usepackage{hyperref}       
\usepackage{url}            
\usepackage{booktabs}       
\usepackage{amsfonts}       
\usepackage{nicefrac}       
\usepackage{microtype}      
\usepackage{xcolor}         
\usepackage{amsmath}
\usepackage{amsthm}
\usepackage{amssymb}
\usepackage{mathtools}
\usepackage{algorithmic}
\usepackage{algorithm}
\usepackage{wrapfig}
\theoremstyle{plain}
\newtheorem{thm}{Theorem}[section]

\newtheorem{lem}[thm]{Lemma}

\theoremstyle{definition}
\newtheorem{definition}[thm]{Definition}
\newtheorem{assumption}{Assumption}
\theoremstyle{remark}
\newtheorem{rem}[thm]{Remark}

\usepackage[round]{natbib}

\begin{document}
\runningtitle{Squeeze All: Novel Estimator and Self-Normalized Bound for Linear Contextual Bandits}


\global\long\def\Expectation{\mathbb{E}}%
\global\long\def\Probability{\mathbb{P}}%
\global\long\def\Var{\mathbb{V}}%
\global\long\def\CE#1#2{\Expectation\left[\left.#1\right|#2\right]}%
\global\long\def\CP#1#2{\Probability\left(\left.#1\right|#2\right)}%
\global\long\def\Real{\mathbb{R}}%
\global\long\def\abs#1{\left|#1\right|}%
\global\long\def\norm#1{\left\Vert #1\right\Vert }%
\global\long\def\Indicator#1{\mathbb{I}\left(#1\right)}%
\global\long\def\Regret#1{\texttt{regret}\ensuremath{(#1)}}%
\global\long\def\Psuedoregret#1{\tilde{\texttt{regret}}\ensuremath{(#1)}}%
\global\long\def\Maxresidual#1#2{\Delta_{#2}\left(#1\right)}%
\global\long\def\Optimalarm#1{a_{#1}^{*}}%
\global\long\def\Action#1{a_{#1}}%
\global\long\def\Tildeaction#1{h_{#1}}%
\global\long\def\Filtration#1{\mathcal{F}_{#1}}%
\global\long\def\History#1{\mathcal{H}_{#1}}%
\global\long\def\Context#1#2{X_{#1,#2}}%
\global\long\def\XX#1#2{\boldsymbol{X}_{#1,#2}}%
\global\long\def\Reward#1{Y_{#1}}%
\global\long\def\Tildereward#1#2{\tilde{Y}_{#1,#2}}%
\global\long\def\TildeError#1#2{\tilde{\eta}_{#1,#2}}%
\global\long\def\SelectionP#1#2{\pi_{#1,#2}}%
\global\long\def\Impute#1{\check{\beta}_{#1}}%
\global\long\def\Estimator#1{\widehat{\beta}_{#1}}%
\global\long\def\Error#1#2{\eta_{#1,#2}}%
\global\long\def\Maxeigen#1{\lambda_{\max}\left(#1\right)}%
\global\long\def\Mineigen#1{\lambda_{\min}\left(#1\right)}%
\global\long\def\Trace#1{\text{Tr}\left(#1\right)}%
\global\long\def\SetofContexts#1{\mathcal{X}_{#1}}%
\global\long\def\Ridgebeta#1{\widehat{\beta}_{#1}^{ridge}}%

\global\long\def\MCP{\textcolor{red}{}}%
\global\long\def\MHO{\textcolor{blue}{}}%
\global\long\def\WYK{\textcolor{orange}{}}%
 \newcommand*{\argmax}{\mathop{\mathrm{argmax}}}

\twocolumn[

\aistatstitle{Squeeze All: Novel Estimator and Self-Normalized Bound\\for Linear Contextual Bandits}

\aistatsauthor{Wonyoung Kim \And  Myunghee Cho Paik  \And Min-hwan Oh}

\aistatsaddress{Columbia University\And Seoul National University,\\ Shepherd23 Inc.  \And  Seoul National University }
]

\begin{abstract}
We propose a linear contextual bandit algorithm with $O(\sqrt{dT\log T})$
regret bound, where $d$ is the dimension of contexts and $T$ is
the time horizon. Our proposed algorithm is equipped with a novel
estimator in which exploration is embedded through explicit randomization.
Depending on the randomization, our proposed estimator takes contribution
either from contexts of all arms or from selected contexts. We establish
a self-normalized bound for our estimator, which allows a novel decomposition
of the cumulative regret into \textit{additive} dimension-dependent
terms instead of multiplicative terms. We also prove a novel lower
bound of $\Omega(\sqrt{dT})$ under our problem setting. Hence, the
regret of our proposed algorithm matches the lower bound up to logarithmic
factors. The numerical experiments support the theoretical guarantees
and show that our proposed method outperforms the existing linear
bandit algorithms. 
\end{abstract}

\section{INTRODUCTION}

The multi-armed bandit (MAB) is a sequential decision making problem
where a learner repeatedly chooses an arm and receives a reward as
partial feedback associated with the selected arm only. The goal of
the learner is to maximize cumulative rewards over a horizon of length~$T$
by suitably balancing exploitation and exploration. The \textit{Linear
contextual bandit} is a general version of the MAB problem, where
$d$-dimensional context vectors are given for each of the arms and
the expected rewards for each arm is a linear function of the corresponding
context vector.

There are a family of algorithms that utilize the principle of \textit{optimism
in the face of uncertainty} (OFU) \citep{lai1985asymptotically}.
These algorithms for the linear contextual bandit have been widely
used in practice (e.g., news recommendation in \citet{li2010contextual})
and extensively analyzed \citep{auer2002using,dani2008stochastic,rusmevichientong2010linearly,chu2011contextual,abbasi2011improved}.
Some of the most widely used algorithms in this family are \texttt{LinUCB}
\citep{li2010contextual} and \texttt{OFUL} \citep{abbasi2011improved}
due to their practicality and performance guarantees. The best known
regret bound for these algorithms is $\tilde{O}(d\sqrt{T})$, where
$\tilde{O}$ stands for big-$O$ notation up to logarithmic factors
of $T$. Another widely-known family of bandit algorithms are based
on randomized exploration, such as Thompson sampling \citep{1933Thompson}.
\texttt{LinTS} \citep{agrawal2013thompson,abeille2017linear} is a
linear contextual bandit version of Thompson sampling with $\tilde{O}(d^{3/2}\sqrt{T})$
or $\tilde{O}(d\sqrt{T\log N})$ regret bound, where $N$ is the total
number of arms. More recently proposed methods based on random perturbation
of rewards \citep{perturb20akveton} also have the same order of regret
bound as \texttt{LinTS}. Hence, many practical linear contextual bandit
algorithms have linear or super-linear dependence on $d$.

A regret bound with sublinear dependence on $d$ has been shown for
\texttt{SupLinUCB}~\citep{chu2011contextual} with $\tilde{O}(\sqrt{dT}\log^{3/2}N)$
regret as well as a matching lower bound $\Omega(\sqrt{dT})$, hence
provably optimal up to logarithmic factors. A more recently proposed
variant of \texttt{SupLinUCB} has been shown to achieve an improved
regret bound of $\tilde{O}(\sqrt{dT\log N})$ \citep{li2019nearly}.
\texttt{SupLinUCB} and its variants (e.g., \citealt{li2017provably,li2019nearly})
improve the regret bound by $\sqrt{d}$ factor capitalizing on independence
of samples via a phased bandit technique proposed by \citet{auer2002using}.
Despite their provable near-optimality, all the algorithms based on
the framework of \citet{auer2002using} including \texttt{SupLinUCB}
tend to explore excessively with insufficient adaptation and are not
practically attractive due to computational inefficiency. Moreover,
the question of whether $\tilde{O}(\sqrt{dT})$ regret is attainable
without relying on the framework of \citet{auer2002using} has remained
open.

A tighter regret bound of \texttt{SupLinUCB} and its variants than
that of \texttt{LinUCB} (and \texttt{OFUL}) stems from utilizing phases
by handling computation separately for each phase. In phased algorithms such as \texttt{SupLinUCB},
the arms in the same phase are chosen without making use of the rewards
in the same phase. This independence of samples allows to apply a
tight confidence bound, improving the regret bound by $\sqrt{d}$
factor. 
On the other hand, this operation should be handled for each
arm, which costs polylogarithmic dependence on $N$ by invoking the
union bound over the arms at the expense of improving $\sqrt{d}$.
In non-phased algorithms such as \texttt{LinUCB} and \texttt{LinTS},
the estimate is adaptive in a sense that the update is made in every
round using all samples collected up to each round; hence the independence
argument cannot be utilized. For this, the well-known self-normalized
theorem \citep{abbasi2011improved} helps avoid the dependence on
$N$, however incurring a linear dependence on $d$ (or super-linear
dependence for \texttt{LinTS}). Thus, the following fundamental question
remains open:

\textit{Can we design a linear contextual bandit algorithm that achieves
a sublinear dependence on $d$ and is adaptive?}

To this end, we propose a novel contextual bandit algorithm that enjoys
the best of the both worlds, achieving a faster rate of $O(\sqrt{dT\log T})$
regret and utilizing adaptive estimation which overcomes the impracticality
of the existing phased algorithms. The established regret bound of our algorithm
matches the regret bound of \texttt{SupLinUCB} in terms of $d$ without
resorting to independence and improves upon it in that its main order
does not depend on $N$. The proposed algorithm is equipped with a
novel estimator in which exploration is embedded through explicit
randomization. Depending on the randomization, the novel estimator
takes contribution either from full contexts or from selected contexts.
Using full contexts is essential in overcoming the dependence due
to adaptivity. Explicit randomization has dual roles. First, the randomization
allows constructing pseudo-outcomes in in~\eqref{eq:pseudo_reward} and thus including all contexts
along with \eqref{eq:pseudo_reward}. Second, randomization promotes the level of exploration
by introducing external uncertainty in the estimator that can be deterministically
computed given observed data. These two features allow a novel additive
decomposition of the regret which can be bounded using the self-normalized
norm of the proposed estimator.

Our main contributions are as follows:
\begin{itemize}
\item We propose a novel algorithm, \textit{Hybridization by Randomization}
bandit algorithm (\texttt{HyRan Bandit}) for a linear contextual bandit.
Our proposed algorithm has two notable features: the first is to utilize
the contexts of all arms both selected and unselected for parameter
estimation, and the second is to randomly perturb the contribution
to the estimator.
\item We establish that our proposed algorithm, \texttt{HyRan Bandit}, achieves
$O(\sqrt{dT\log T})$ regret upper bound without dependence on $N$
on the leading term.
Ours is the first method achieving $\tilde{O} (\sqrt{dT})$ regret
without relying on the widely used technique by \citet{auer2002using}
and its variants (e.g., \texttt{SupLinUCB}). To the best of our knowledge,
this is the fastest rate regret bound for the linear contextual bandit.
\item We propose a novel \texttt{HyRan} (Hybridization by randomization)
estimator which uses either the contexts of all arms or selected contexts
depending on randomization. We establish a self-normalized bound (Theorem~\ref{thm:Estimation_error})
for our estimator, which allows a novel decomposition of the cumulative
regret into \textit{additive} dimension-dependent terms (Lemma~\ref{lem:regret_decomposition})
instead of multiplicative terms. This allows us to establish the faster
rate of the cumulative regret.
\item We prove a novel lower bound of $\Omega(\sqrt{dT})$ for the cumulative
regrets (Theorem~\ref{thm:lower_bound}) under our problem setting.
The lower bound matches with the regret upper bound of \texttt{HyRan
Bandit} up to logarithmic factors, hence showing the provable near-optimality
of our method.
\item We evaluate \texttt{HyRan Bandit} on numerical experiments and show
that the practical performance of our proposed algorithm is in line
with the theoretical guarantees and is superior to the existing algorithms. 
\end{itemize}


\section{RELATED WORKS}

The linear contextual bandit problem was first introduced by \citet{abe1999associative}.
UCB algorithms for the linear contextual bandit have been proposed
and analyzed by \citet{auer2002using,dani2008stochastic,rusmevichientong2010linearly,chu2011contextual,abbasi2011improved}
and their follow-up works. Thompson sampling based algorithms have
also been widely studied \citep{agrawal2013thompson,abeille2017linear}.
Both classes of the algorithms typically have linear (or superlinear)
dependence on context dimension. To our knowledge, all of the regret bounds with sublinear dependence on context dimension are for UCB
algorithms based on the IID sample generation technique of \citet{auer2002using}.
The examples include \texttt{SupLinUCB} \citet{chu2011contextual}
with an $O\big(\sqrt{dT}\log^{3/2}(NT)\big)$ regret bound and its variant \texttt{VCL-SupLinUCB} \citep{li2019nearly} with an $O(\sqrt{dT(\log T)(\log N)})\cdot\text{poly}(\log\log(NT))$ regret bound. 
The phase-based elimination algorithms with $O(\sqrt{dT\log NT})$ regret bound introduced by \citet{valko2014spectral} and \citet{lattimore2020bandit} is a variant of \texttt{SupLinUCB} for the case where the set of contexts does not change over time.
Despite their sharp regret bounds, these \texttt{SupLinUCB}-type
algorithms based on the framework of \citet{auer2002using} are impractical due to its algorithmic design to discard the observed rewards and to explore excessively with insufficient adaptation.

The rewards for the unselected arms are not observed, hence, missing.
Recently some bandit literature has framed the bandit setting as a
missing data problem, and employed missing data methodologies \citep{dimakopoulou2019balanced,kim2019doubly,kim2021doubly}.
\citet{dimakopoulou2019balanced} employs an \textit{inverse probability
weighting} (IPW) estimator using the selected contexts alone and proves
an $\tilde{O}(d\sqrt{\epsilon^{-1}T^{1+\epsilon}N})$ regret bound
for \texttt{LinTS} which depends on the number of arms, $N$. The
\textit{doubly robust} (DR) method \citep{robins1994,bang2005doubly}
is adopted in \citet{kim2019doubly} with Lasso penalty for high-dimensional
settings with sparsity and the regret bound is shown to be improved
in terms of the sparse dimension instead of $d$. Recently in \citet{kim2021doubly},
a modified \texttt{LinTS} employing the DR method is proposed and
provided an $\tilde{O}(d\sqrt{T})$ regret bound. The authors improve
the bound by using contexts of all arms including the unselected ones
which paves a way to circumvent the technical definition of unsaturated
arms.

A key element in building the DR method is a random variable with
a known probability distribution. In Thompson sampling, randomness
is inherent in the step sampling from a posterior distribution, and
the probability of the selected arm having the largest predicted outcome
can be computed. This allows naturally constructing the DR estimator.
All previous DR-type estimators capitalize on randomness in Thompson
sampling (e.g. \citet{dimakopoulou2019balanced,kim2021doubly}) or
in epsilon-greedy \citep{kim2019doubly}. In algorithms without such
inherent randomness, the DR estimators cannot be constructed. In this
paper, we generate a random variable to determine whether to contribute
full contexts or just chosen context.

Another line of the literature that uses stochastic assumptions on
contexts include \citet{goldenshluger2013linear,bastani2020online}
and \citet{bastani2021mostly}. In their work, the problem setting
is different from ours in that they consider $N$ different parameters
for each arm with single context vector shared for all arms. They
resort to much stronger assumptions for regret analysis such as the
margin condition \citep{goldenshluger2013linear,bastani2020online,bastani2021mostly}
as well as the covariate diversity condition \citep{bastani2021mostly}
that allow for a greedy approach to be efficient. However, in our
problem setting, such assumptions are not applied and a simple greedy
policy would cause regret linear in $T$.



\section{LINEAR CONTEXTUAL BANDIT PROBLEM}

In each round $t\in[T]:=\{1,\ldots,T\}$, the learner observes a set
of arms $[N]:=\{1,...,N\}$ with their corresponding context vectors
$\{\Context it\in\Real^{d}\mid i\in[N]\}$. Then, the learner chooses
an arm $\Action t\in[N]$ and receives a random reward $\Reward t:=Y_{a_{t},t}$
for the chosen arm. For all $t\in[T]$ and $i\in[N]$, we assume the
linear reward model, i.e., $Y_{i,t}=\Context it^{T}\beta^{*}+\Error it$,
where $\beta^{*}\in\Real^{d}$ is an \textit{unknown} parameter and
$\Error it\in\Real$ is an independent noise. Let $\History t$ be
the history at round $t$ that contains contexts $\{\Context i{\tau}\}_{i=1,\tau=1}^{N,t}$,
chosen arms $\{\Action{\tau}\}_{\tau=1}^{t-1}$ and the corresponding
rewards $\{\Reward{\tau}\}_{\tau=1}^{t-1}$. For each $t$ and $i$,
the noise $\eta_{i,t}$ is zero-mean conditioned on $\History t$,
i.e, $\CE{\eta_{i,t}}{\History t}=0$. The optimal arm at round $t$
is defined as $\Optimalarm t:=\arg\max_{i\in[N]}\left\{ \Context it^{T}\beta\right\} $.
Let $\Regret t$ be the difference between the expected rewards of
the chosen arm and the optimal arm at round $t$, i.e., $\Regret t:=\Context{\Optimalarm t}t^{T}\beta^{*}-\Context{\Action t}t^{T}\beta^{*}.$
The goal is to minimize the sum of regrets over $T$ rounds, $R(T):=\sum_{t=1}^{T}\Regret t$.
The time horizon $T$ is finite but possibly unknown.


\section{PROPOSED METHODS}

In this section, we present the methodological contributions, the
new estimator (Section~\ref{sec:estimator}) and the new contextual
bandit algorithm that utilizes the proposed estimator (Section~\ref{sec:algorithm}).

\subsection{Hybridization by Randomization (\texttt{HyRan}) Estimator}

\label{sec:estimator}

We start from two candidate estimators, the ridge estimator and the
DR estimator, and their corresponding estimating equations. The first
one, the \textit{ridge score} function is a sum of contribution from
round $\tau$, 
\begin{equation}
\Context{\Action{\tau}}{\tau}\left(Y_{\Action{\tau},\tau}-\Context{\Action{\tau}}{\tau}^{T}\beta\right).\label{eq:ridge_score}
\end{equation}
The other is the \textit{DR score} function. However, to employ the
DR technique in general cases, we need preliminary works. The DR procedure
is originally proposed for missing data problems, and requires the
observation (or missing) indicator and the observation probability
as the main elements. These two elements are naturally provided in
Thompson sampling: the indicator $a_{t}$ being each arm is a random
variable given history since the estimator is sampled from a posterior
distribution, and the expectation of this indicator is the probability
of choosing each arm. All previous DR-typed bandits apply the DR technique
to algorithms equipped with inherent randomness such as Thompson sampling
or epsilon-greedy. The DR procedure cannot be naturally applied to
the algorithms without inherent randomness, e.g. \texttt{LinUCB},
since the indicator that $a_{t}$ equals each arm is not random but
deterministic given history. For the DR technique to be applied regardless
whether $a_{t}$ is random or not, we introduce an external random
device by sampling $h_{t}$ from $[N]$ with a known non-zero probability.
We can convert $h_{t}$ into $N$-variate one hot vector following
a multinomial distribution. Thanks to this seemingly superfluous external
random variable through $h_{t}$, we can construct a DR score, whose
contribution at round $\tau$ is: 
\begin{equation}
\sum_{i=1}^{N}\Context i{\tau}\left(\Tildereward i{\tau}-\Context i{\tau}^{T}\beta\right),\label{eq:DR_score}
\end{equation}
where the pseudo reward $\Tildereward i{\tau}$ is defined as 
\begin{equation}
\Tildereward i{\tau}=\left\{ 1-\frac{\Indicator{\Tildeaction{\tau}=i}}{\SelectionP i{\tau}}\right\} \Context i{\tau}^{T}\Impute{\tau}+\frac{\Indicator{\Tildeaction{\tau}=i}}{\SelectionP i{\tau}}Y_{\Tildeaction{\tau},\tau},\label{eq:pseudo_reward}
\end{equation}
for some random variable $\Tildeaction{\tau}$ sampled from $[N]$,
with probability $\SelectionP i{\tau}:=\Probability(\Tildeaction{\tau}=i)$,
and $\Impute{\tau}$ is an imputation estimator defined in Section~\ref{subsec:imputation_estimator}.
The DR score~\eqref{eq:DR_score} uses $\Tildereward i{\tau}$ instead
of $Y_{i,\tau}$ in the original score function to estimate $\beta$
as if all rewards were observed. Using the pseudo reward~\eqref{eq:pseudo_reward},
we can use all contexts rather than just selected contexts.

Although the external random variable paves a way to utilize DR techniques,
it also causes trouble in computing~\eqref{eq:pseudo_reward} since
$Y_{i,t}$ is observed for $i=a_{t}$ not for $i=h_{t}$. Therefore
the second term of~\eqref{eq:pseudo_reward} cannot be computed if
$h_{t}\neq a_{t}$. The solution to this problem shapes the main theme
of our proposed method, namely \textit{hybridization}. Our strategy
is to construct a score function from~\eqref{eq:DR_score} when $h_{t}=a_{t}$,
but from~\eqref{eq:ridge_score} when $h_{t}\neq a_{t}.$

We denote the indices of $t$ by $\Psi_{t}$ if $\Tildeaction t=\Action t$.
With the subsampled set of rounds $\Psi_{t}$ we can define our hybrid
score equation 
\begin{equation}
\begin{split} & \sum_{\tau\in\Psi_{t}}\sum_{i=1}^{N}\Context i{\tau}\left(\Tildereward i{\tau}-\Context i{\tau}^{T}\beta\right)\\
 & +\sum_{\tau\notin\Psi_{t}}\Context{\Action{\tau}}{\tau}\left(Y_{\Action{\tau},\tau}-\Context{\Action{\tau}}{\tau}^{T}\beta\right)+\lambda_{t}\beta=0.
\end{split}
\label{eq:hybrid_score}
\end{equation}
The first term is from the DR score~\eqref{eq:DR_score} and the
second term is from the ridge score~\eqref{eq:ridge_score}. The
contribution of the two score functions is determined by the subset
$\Psi_{t}$ which is randomized with the random variable $\Tildeaction t$.
Therefore, we call the random variable $\Tildeaction t$ as a \textit{hybridization
variable}. Specifically, for each round $t\in[T]$ and given $p\in(0,1)$,
we sample $\Tildeaction t$ from $[N]$ with probability, 
\begin{equation}
\begin{split}\SelectionP{\Action t}t & :=\CP{\Tildeaction t=\Action t}{\Filtration t}=p,\\
\SelectionP jt & :=\CP{\Tildeaction t=j}{\Filtration t}=\frac{1-p}{N-1},\;\forall j\neq\Action t,
\end{split}
\label{eq:tilde_prob}
\end{equation}
where $\Filtration t:=\History t\cup\{\Action t\}\cup\{\Tildeaction 1,\ldots,\Tildeaction{t-1}\}$.
We emphasize that $\Tildeaction t$ is sampled after an arm $\Action t$
is pulled and does not affect the choice of $\Action t$. 

Our proposed estimator is the solution of~\eqref{eq:hybrid_score}
which can be written as 
\begin{equation}
\begin{split}\Estimator t\!:= & \!\left(\!\sum_{\tau\in\Psi_{t}}\sum_{i=1}^{N}\Context i{\tau}\Context i{\tau}^{T}\!+\!\!\sum_{\tau\notin\Psi_{t}}\Context{\Action{\tau}}{\tau}\Context{\Action{\tau}}{\tau}^{T}\!+\!\lambda_{t}I\!\right)^{-1}\\
 & \left(\sum_{\tau\in\Psi_{t}}\sum_{i=1}^{N}\Context i{\tau}\Tildereward i{\tau}\!+\!\!\sum_{\tau\notin\Psi_{t}}\Context{\Action{\tau}}{\tau}\Reward{\tau}\right).
\end{split}
\label{eq:estimator}
\end{equation}
This is a hybrid form of using the contexts of all arms and using
the contexts of the selected arms, and the contribution is set by
the random variable the subsampled rounds $\Psi_{t}$. We later provide
the estimation error bound for this newly proposed estimator in Theorem~\ref{thm:Estimation_error}
which allows us to shave off the dimensionality dependence in regret
analysis.


\begin{algorithm}[t]
\caption{Hybridization by Randomization Bandit Algorithm for Linear Contextual Bandits}
\label{alg:HyRan} 
\begin{algorithmic} 
\STATE \textbf{INPUT}:
Regularization parameter $\lambda_{t}>0$, subsampling parameter $p\in(0,1)$.
\STATE Initialize $V_{0}=I_{d}$, $Z_{0}=0_d$ 
\FOR{$t=1$ to $T$} 
\STATE Observe contexts $\{\Context it\}_{i=1}^{N}$ and estimate $\Estimator{t-1}=\left(V_{t-1}+\lambda_{t}I_{d}\right)^{-1}Z_{t-1}$
\STATE Play $\Action t=\arg\max_{i}\Context it^{T}\Estimator{t-1}$ and observe $\Reward t$ \STATE Set $\SelectionP{\Action t}t:=p$
and $\SelectionP jt:=\frac{1-p}{N-1}$ for $j\neq\Action t$ 
\STATE Sample a hybridization variable $\Tildeaction t$ from the multinomial
distribution with probability ($\SelectionP 1t,...,\SelectionP Nt$)
\IF{$\Tildeaction t=\Action t$} 
\STATE Update $V_{t}=V_{t-1}+\sum_{i=1}^{N}\Context it\Context it^{T}$
and $Z_{t}=Z_{t-1}+\sum_{i=1}^{N}\Context it\Tildereward it$ \ELSE
\STATE Update $V_{t}=V_{t-1}+\Context{\Action t}t\Context{\Action t}t^{T}$
and $Z_{t}=Z_{t-1}+\Context{\Action t}t\Reward t$ 
\ENDIF 
\STATE Update $\Impute t=\left(V_{t}+\sqrt{t}I_{d}\right)^{-1}Z_{t}$ 
\ENDFOR
\end{algorithmic}
\end{algorithm}

\subsection{\texttt{HyRan Bandit} Algorithm}

\label{sec:algorithm}

\begin{figure}[t]
\centering \includegraphics[width=0.41\textwidth]{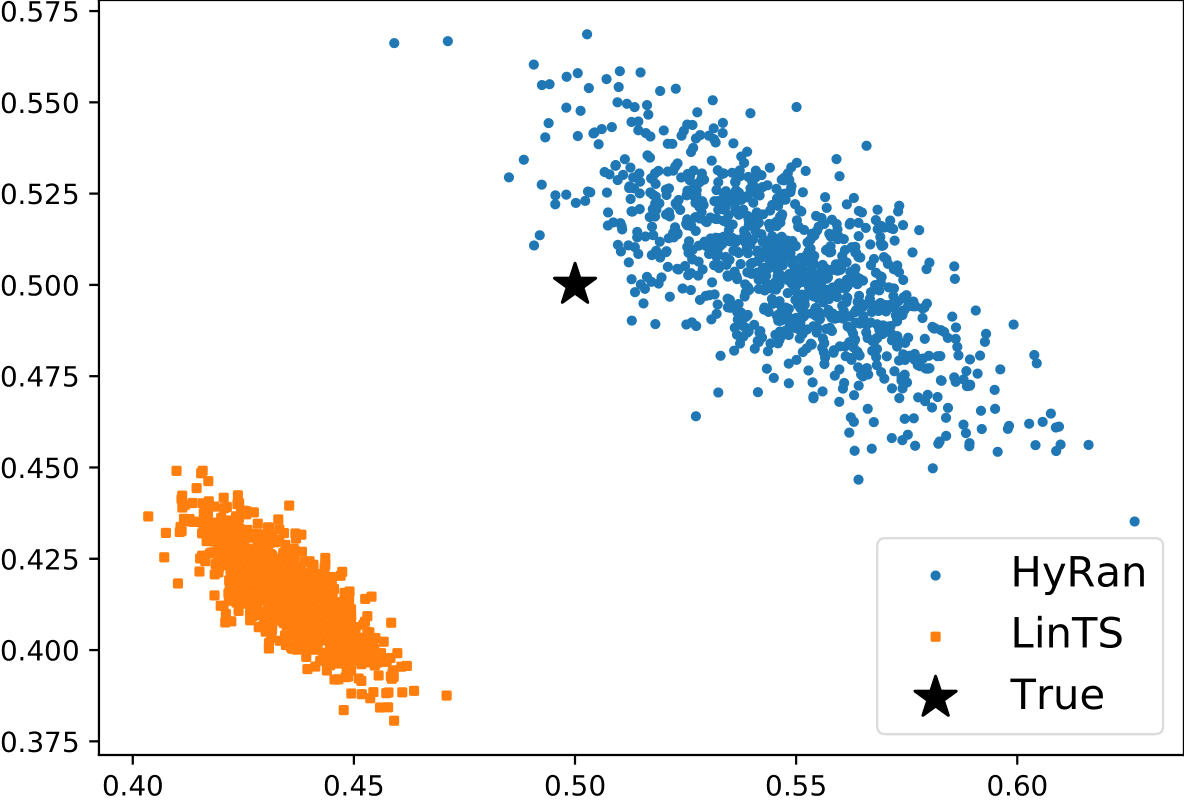}
\caption{ An illustration of the 1000 generated estimators of $\beta^{*}$
used in \texttt{HyRan Bandit} and \texttt{LinTS} at round $t=1000$,
when $d=2$ and $N=5$. The points in blue and orange represent the
generated \texttt{HyRan} and \texttt{LinTS} estimators, respectively.
The black star in the plot represents the true parameter~$\beta^{*}$. }
\label{fig:contours} 
\end{figure}

Our proposed algorithm, \texttt{HyRan Bandit}, is presented in Algorithm~\ref{alg:HyRan}.
At each round~$t$, the algorithm computes $\Context it^{T}\Estimator{t-1}$
for each arm $i\in[N]$ based on our estimator~(\ref{eq:estimator})
and finds the arm $\Action t$ with the maximum estimated reward.
After pulling $\Action t$ and observing the reward for the selected
arm, the next step is to determine whether the contribution to the
estimator is the ridge score~\eqref{eq:ridge_score} or the DR score~\eqref{eq:DR_score}.
\texttt{HyRan Bandit} then samples the hybridization variable $\Tildeaction t\in[N]$
from the multinomial distribution with probability ($\SelectionP 1t,...,\SelectionP Nt$).
This procedure determines whether the contexts and reward at round
$t$ is added by ~\eqref{eq:ridge_score} or ~\eqref{eq:DR_score}.
When $\Tildeaction t$ is equal to $\Action t$, we can observe the
reward $\Tildereward{\Tildeaction t}t$ and compute the pseudo reward
in \eqref{eq:pseudo_reward}. Therefore we include the round $t$
in $\Psi_{t}$, and use~\eqref{eq:DR_score}, otherwise we use ~\eqref{eq:ridge_score}.
When the contribution to the score function is determined, \texttt{HyRan
Bandit} updates $\Estimator t$ as in~\eqref{eq:estimator}.

In order to compute $\Estimator t$, the algorithm requires another
imputation estimator $\Impute t$ to determine the pseudo reward in
\eqref{eq:pseudo_reward}. In order to obtain the near-optimal regret
bound, one must use an imputation estimator such that $\norm{\Impute t-\beta^{*}}_{2}\le N^{-1}$
holds after some explorations. 
For the definition of the imputation estimator $\Impute t$ used in
our analysis, see Section \ref{subsec:imputation_estimator}. Since
$\Impute t$ is multiplied with mean zero random variable in \eqref{eq:pseudo_reward}
the unbiasedness of the estimator does not depend on the choice of
$\Impute t$.

\textit{Discussion of the algorithm.} The action selection in \texttt{HyRan
Bandit} is greedy given the \texttt{HyRan} estimator. However the
algorithm is not exploration-free since the \texttt{HyRan} estimator
is generated randomly. Note that action selection in \texttt{LinTS}
is also greedy given the sampled estimator. The estimator from \texttt{LinTS}
represents a realization from a posterior distribution. Hence, exploration
is embedded in the estimator through variability in the distribution.
Similarly, in our method, exploration is embedded in the \texttt{HyRan}
estimator. Our estimator represents a realization of random variables
corresponding to a particular subset $\Psi_{t}$ out of all possible
subsets. Therefore, exploration is inherent from the variability of
randomization scheme. For the sake of illustrating inherent exploration,
we purposely generate multiple estimators both for \texttt{HyRan}
and \texttt{LinTS} in a given round. Note that both algorithms compute
only a single estimator per round. In Figure~\ref{fig:contours},
the points in blue represent the \texttt{HyRan} estimators of $\beta^{*}$
from many possible realizations of $\Psi_{t}$ due to the randomness
of $h_{t}$. For \texttt{LinTS}, the points in orange represent the
sampled estimators of $\beta^{*}$ from its posterior distribution.
We observe that there is enough variability for our estimator as in
\texttt{LinTS}.


\section{MAIN RESULTS}

\label{sec:results}

In this section, we present our main theoretical results: the regret
bound for \texttt{HyRan Bandit} (Theorem~\ref{thm:regret_bound})
and the estimation error bound of the proposed 
\texttt{HyRan} estimator (Theorem~\ref{thm:Estimation_error}). We
first provide the assumptions used throughout the analysis.

\begin{assumption}[Boundedness]\label{assum:boundedness} For all
$i\in[N]$ and $t\in[T]$, $\norm{\Context it}_{2}\le1$ and $\norm{\beta^{*}}_{2}\le1$.
\end{assumption}

\begin{assumption}[Sub-Gaussian noise]\label{assum:noise} For
each $t$ and $i$, the noise $\Error it$ is conditionally $\sigma$-sub-Gaussian
for a fixed constant $\sigma\ge0$, i.e, 
$\CE{\exp\left(\lambda\eta_{i,t}\right)}{\History t}\le\exp(\lambda^{2}\sigma^{2}/2)$,
for all $\lambda\in\Real$. \end{assumption}



\begin{assumption}[Context stochasticity] \label{assum:iid_context}
The set of context vectors $\mathcal{X}_{t}:=\{\Context it\in\Real^{d}:i\in[N]\}$
is independently drawn from unknown distribution $P_{\mathcal{X}}$
with $\lambda_{\min}(\Expectation[\frac{1}{N}\sum_{i=1}^{N}\Context it\Context it^{T}])\ge\phi^{2}>0$,
for all $t$. \end{assumption}

\textit{Discussion of the assumptions.} Assumptions~\ref{assum:boundedness}
and \ref{assum:noise} are standard in the stochastic contextual bandit
literature (see e.g. \citet{agrawal2013thompson}). The same or similar
assumption to Assumption~\ref{assum:iid_context} has been frequently
used in the contextual bandit literature \citep{goldenshluger2013linear,li2017provably,bastani2020online,oh2021sparsity,kim2021doubly}.
We emphasize that stochasticity is assumed for the entire context
set and that we allow context vectors to be correlated in each round.
We also emphasize that even under the stochasticity of contexts, achieving
a regret bound sublinear in $d$ was only possible by resorting to
the technique as used in \texttt{SupLinUCB} \citep{auer2002using}
and its follow-up works.

The positive-definiteness on the average of the covariance matrix
in Assumption \ref{assum:iid_context} can be satisfied regardless
of the number of arms - even when N = 1, e.g., when the context vector
(s) is (are) drawn from the Uniform distribution or the truncated
Gaussian distribution. Recently, \citet{bastani2021mostly,kim2022double}
identified the practical cases where Assumption \ref{assum:iid_context}
holds. Technically, Assumption \ref{assum:iid_context} is required
to obtain the fast convergence rate in estimating linearly parametrized
responses in Statistics (see e.g., \citet{buhlmann2011statistics}).
In our work the assumption is used to obtain the fast convergence
rate for the imputation estimator (Lemma B.3).




\subsection{Regret Bound of \texttt{HyRan Bandit}}

\label{subsec:regret} Under the assumptions above, we present the
following regret bound for the \texttt{HyRan Bandit} algorithm.
\begin{thm}
Suppose Assumptions~1-3 hold and the total number of rounds $T$
satisfies 
\begin{equation}
T\ge\mathcal{E}=\max\left\{ \frac{8}{p}\log\frac{T}{\delta},C_{p,\sigma}N^{2}\phi^{-4}\log\frac{2T}{\delta}\right\} ,\label{eq:exploration_term}
\end{equation}
where $C_{p,\sigma}:=\frac{8(2-p)}{(1-p)\sqrt{p}}+\frac{\sqrt{2}C\sigma}{p^{2}}+\frac{8}{\sqrt{p}}$
is a constant depending only on $p$ and $\sigma$. Set $\lambda_{t}:=d\log\frac{4t^{2}}{\delta}$.
Then the total regret by time $T$ for \texttt{HyRan Bandit} is bounded
by 
\begin{equation}
\begin{split}R(T)\le & 2\mathcal{E}+4D_{p,\sigma}\sqrt{2T\log\frac{1}{\delta}}+3\delta D_{p,\sigma}\\
 & +\frac{\left(16\sqrt{2}+8\right)D_{p,\sigma}}{\sqrt{p}}\sqrt{dT\log\frac{2T}{\delta}},
\end{split}
\label{eq:regret_bound}
\end{equation}
with probability at least $1-8\delta$, where $D_{p,\sigma}:=1+\frac{4\sqrt{2}}{1-p}+\frac{\sigma}{p}$
is a constant depending only on $p$ and $\sigma$. \label{thm:regret_bound} 
\end{thm}

\textit{Discussion on the regret bound.} The subsampling parameter
$p\in(0,1)$ in \texttt{HyRan Bandit} is chosen independently with
respect to $N$, $d$ or $T$ and does not affect the rate of our
regret bound. The number of rounds $\mathcal{E}$ defined in~\eqref{eq:exploration_term}
is required for the imputation estimator $\Impute t$ to obtain a
suitable estimation error bound which is crucial to derive our self-normalized
bound for \texttt{HyRan} estimator. The number of exploration rounds
is $O(N^{2}\phi^{-4}\log T)$ which is only logarithmic in $T$ and
is bounded by $O(\sqrt{dT\log T})$ when $\frac{T}{\log T} \ge N^{4}d^{-1}\phi^{-8}$.
The value of $\phi^{-2}$ is $O(d)$ for many standard context distributions (see e.g., Lemma 5.2 in \citet{kim2022double}).
As a result, the regret bound of \texttt{HyRan Bandit} is $O(\sqrt{dT\log T})$.
Our bound is sharper than the existing regret bounds of $O(\sqrt{dT\log T\log N})\cdot\text{poly}(\log\log(NT))$
for \texttt{VCL-SupLinUCB} \citep{li2019nearly} and $O(\sqrt{dT}\log^{3/2}(NT))$
for \texttt{SupLinUCB} 
\citep{chu2011contextual}, although direct comparison is not immediate
due to difference in the assumptions used. It is important to note
that the leading term in our regret bound does not depend on $N$
while the existing $\tilde{O}(\sqrt{dT})$ regret bounds all contain
$N$ dependence in their leading terms. 
To our knowledge, the regret bound in Theorem~\ref{thm:regret_bound}
is the fastest rate among linear contextual bandit algorithms. Furthermore,
we believe that \texttt{HyRan Bandit} is the first method achieving
a regret that is sublinear in context dimension \textit{without} using
the widely used technique by \citet{auer2002using} and its variants
(e.g., \texttt{SupLinUCB}).

Our regret bound~in Theorem~\ref{thm:regret_bound} is smaller than
the existing lower bounds for the linear contextual bandits in \citet{rusmevichientong2010linearly,lattimore2020bandit}
and \citet{li2019nearly}. This is not a contradiction since the slightly
different set of assumptions are used. i.e., Assumptions~\ref{assum:iid_context}.
We discuss this issue in Section \ref{subsec:lower_bound} by proving
a lower bound under Assumption~\ref{assum:iid_context}, which matches
with~\eqref{eq:regret_bound} up to a logarithmic factor.

\subsection{Regret Decomposition}

\label{subsec:decomposition}

In the analysis of \texttt{LinUCB} and \texttt{OFUL}, an instantaneous
regret is controlled by using the joint maximizer of the reward 
\[
(a_{t},\widehat{\beta}_{\textrm{ucb}})=\arg\max_{i\in[N],\beta\in\mathcal{C}_{t}}X_{i,t}^{T}\beta
\]
where $\mathcal{C}_{t}$ is a high-probability confidence ellipsoid.
Then, $\Regret t$ is typically decomposed as 
\begin{equation}
\Regret t\le\norm{\widehat{\beta}_{ucb}-\beta^{*}}_{A_{t}}\norm{\Context{\Action t}t}_{A_{t}^{-1}},\label{eq:typical_regret_bound}
\end{equation}
where $A_{t}:=\sum_{\tau=1}^{t}\Context{\Action{\tau}}{\tau}\Context{\Action{\tau}}{\tau}^{T}+\lambda I$.
Each of the two terms on the right hand side in (\ref{eq:typical_regret_bound})
has a $\sqrt{d}$ factor. In particular, $\sqrt{d}$ factor in the
first term comes from the radius of $\mathcal{C}_{t}$. Hence, this
results in $O(d)$ regret when combined.

In our work, we introduce new decomposition of regret that allows
to avoid multiplicative terms. This decomposition allows for non-OFU
based analysis for sharper dependence on dimensionality.

\begin{lem}[Regret decomposition]
\label{lem:regret_decomposition} Define the max-residual function
for $x=(x_{1},\ldots,x_{N})\in\Real^{d\times N}$ as $\Maxresidual x{\widehat{\beta}}:=\max_{i\in[N]}|x_{i}^{T}(\widehat{\beta}-\beta^{*})|\,.$
For each $t\in[T]$, let $\SetofContexts t:=(\Context 1t,\ldots,\Context Nt)$
and denote $\mathcal{G}_{t}:=\cup_{\tau=1}^{t}\{\SetofContexts{\tau},\Estimator{\tau}\}$.
Then for $t\ge1$, 
\begin{equation}
\begin{split} & \Regret{t+1}\\
 & \le2\left\{ \Maxresidual{\SetofContexts{t+1}}{\Estimator t}-\CE{\Maxresidual{\SetofContexts{t+1}}{\Estimator t}}{\mathcal{G}_{t}}\right\} \\
 & +2\left\{ \CE{\Maxresidual{\SetofContexts{t+1}}{\Estimator t}}{\mathcal{G}_{t}}-\frac{1}{\abs{\Psi_{t}}}\sum_{\tau\in\Psi_{t}}\Maxresidual{\SetofContexts{\tau}}{\Estimator t}\right\} \\
 & +\frac{2}{\sqrt{\abs{\Psi_{t}}}}\norm{\Estimator t-\beta^{*}}_{V_{t}},
\end{split}
\label{eq:regret_decomposition}
\end{equation}
where 
\[
V_{t}:=\sum_{\tau\in\Psi_{t}}\sum_{i=1}^{N}\Context i{\tau}\Context i{\tau}^{T}+\sum_{\tau\notin\Psi_{t}}\Context{\Action{\tau}}{\tau}\Context{\Action{\tau}}{\tau}^{T}+\lambda_{t}I.
\]
\end{lem}

The decomposition of the expected regret given in~\eqref{eq:regret_decomposition}
directly bounds the regret by approximating the max-residual with
$t+1$-th contexts $\SetofContexts{t+1}$ to that with the average
over the contexts in round $\tau\in\Psi_{t}$, which is bounded by
the self-normalized bound for \texttt{HyRan} estimator. This approximation
yields two additive terms: the difference between the max-residual
function and its expectation $(\Maxresidual{\SetofContexts{t+1}}{\Estimator t}-\CE{\Maxresidual{\SetofContexts{t+1}}{\Estimator t}}{\mathcal{G}_{t}})$,
and the difference between the expectation over the context distribution
and its empirical distribution ($\CE{\Maxresidual{\SetofContexts{t+1}}{\Estimator t}}{\mathcal{G}_{t}}-\frac{1}{\abs{\Psi_{t}}}\sum_{\tau\in\Psi_{t}}\Maxresidual{\SetofContexts{\tau}}{\Estimator t}$).
The bound becomes tighter as the size of $\Psi_{t}$ increases, because
we can use more contexts for the approximation.

The decomposition is insightful in that the regret from suboptimal
arm selections is incurred due to poor estimate, thus can be bounded
by the quantities involving the maximum residual. To bound the maximum
residual, \texttt{SupLinUCB} and their variants that achieve $\tilde{O}(\sqrt{dT})$
regret bound handle the maximum residual with the union of $N\times T$
probability inequalities, and this gives $\log N$ term in the regret
bound. But in Lemma~\ref{lem:regret_decomposition}, we use the fact
that the maximum residual is bounded by a sum of residuals. The sum
of residuals can be shown to be bounded by the self-normalized bound
for our estimator in~\eqref{eq:estimator}. This replacement is possible
since our novel estimator uses all contexts for some subsampled rounds.
In this way, we can use only $T$ probability inequalities and eliminate
the $N$ independence on the leading term of the regret bound. We
emphasize that the decomposition yields the self-normalized bound
of our new estimator, not any estimator using the contexts of selected
arms only (e.g. ridge estimator for \texttt{OFUL}). Our bound is normalized
with the hybrid Gram matrix $V_{t}$, not that of selected contexts.


To bound the terms in the decomposed instantaneous regret~\eqref{eq:regret_decomposition},
we see that the first term is bounded by using Azuma's inequality.
We bound the second and third term using Lemma~\ref{lem:expectation_bound}
and Theorem \ref{thm:Estimation_error}, respectively. Lemma \ref{lem:expectation_bound}
adopts the empirical theories on the distribution of the contexts.
\begin{lem}
\label{lem:expectation_bound} Suppose Assumptions 1-3 hold. For each
$t\in[T]$, and $L>0$, conditioned on $\Psi_{t}$, with probability
at least $1-\delta/T$, 
\begin{align*}
 & \sup_{\norm{\beta_{1}-\beta^{*}}_{2}\le L}\abs{\CE{\Maxresidual{\SetofContexts{t+1}}{\beta_{1}}}{\mathcal{G}_{t}}-\frac{1}{\abs{\Psi_{t}}}\sum_{\tau\in\Psi_{t}}\Maxresidual{\SetofContexts{\tau}}{\beta_{1}}}\\
 & \le\frac{3L\delta}{2T}+4L\sqrt{\frac{1}{\abs{\Psi_{t}}}}\sqrt{d\log\frac{2T}{\delta}}.
\end{align*}
\end{lem}

In the following theorem, we present the self-normalized bound for
the compound estimator which allows us to bound the last term in (\ref{eq:regret_decomposition}).
\begin{thm}[A self-normalized bound for \texttt{HyRan} estimator]
\label{thm:Estimation_error} Suppose Assumptions \ref{assum:boundedness}-\ref{assum:iid_context}
hold. 
Let $\Estimator t$ be the estimator defined in \eqref{eq:estimator} and  $p\in(0,1)$ be a constant used in \eqref{eq:tilde_prob}. 
Then with probability at least $1-6\delta$, 
\begin{equation}
\norm{\Estimator t-\beta^{*}}_{V_{t}}\le\sqrt{\lambda_{t}}+\left(\frac{4\sqrt{2}}{1-p}+\frac{\sigma}{p}\right)\sqrt{d\log\frac{4t^{2}}{\delta}},\label{eq:estimation_error_bound}
\end{equation}
for all $t\ge\max\left\{ \frac{8}{p}\log\frac{T}{\delta},C_{p,\sigma}N^{2}\phi^{-4}\log\frac{2T}{\delta}\right\} $,
where $C_{p,\sigma}>0$ is a constant depending only on $p$ and $\sigma$. 
\end{thm}

\begin{figure*}[t]
\centering{}\includegraphics[width=0.32\linewidth]{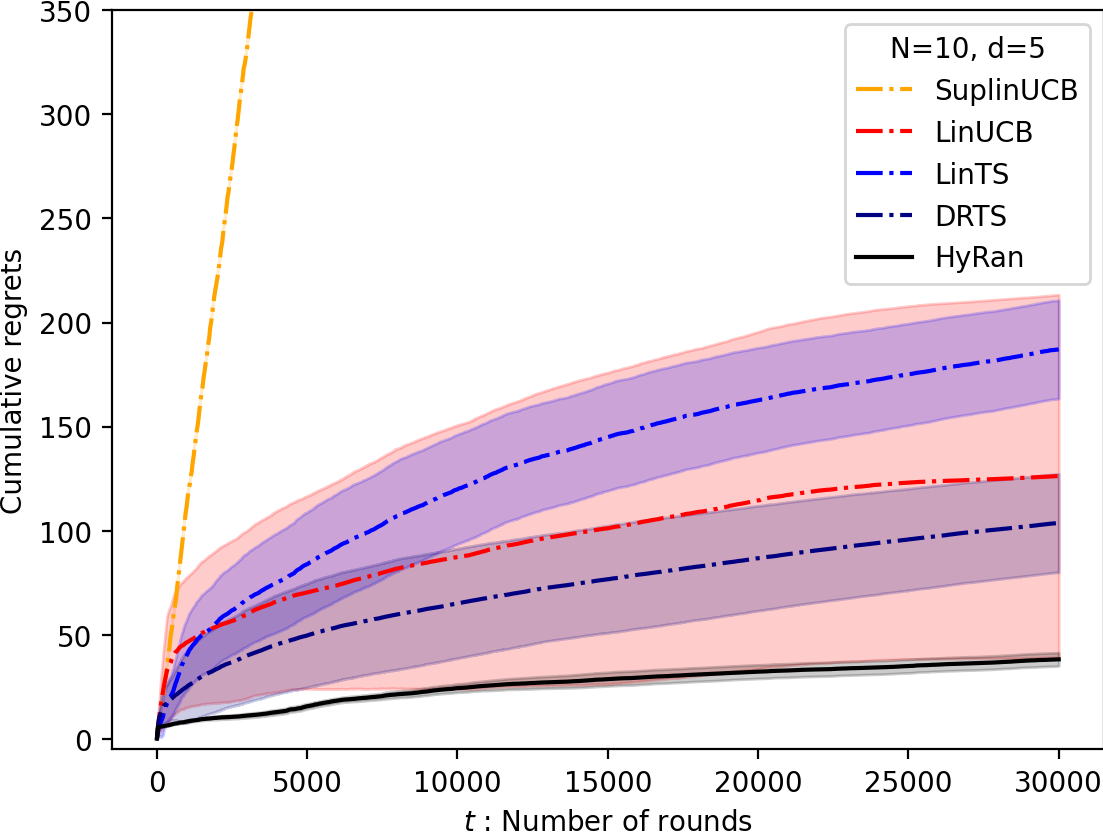}
\includegraphics[width=0.32\linewidth]{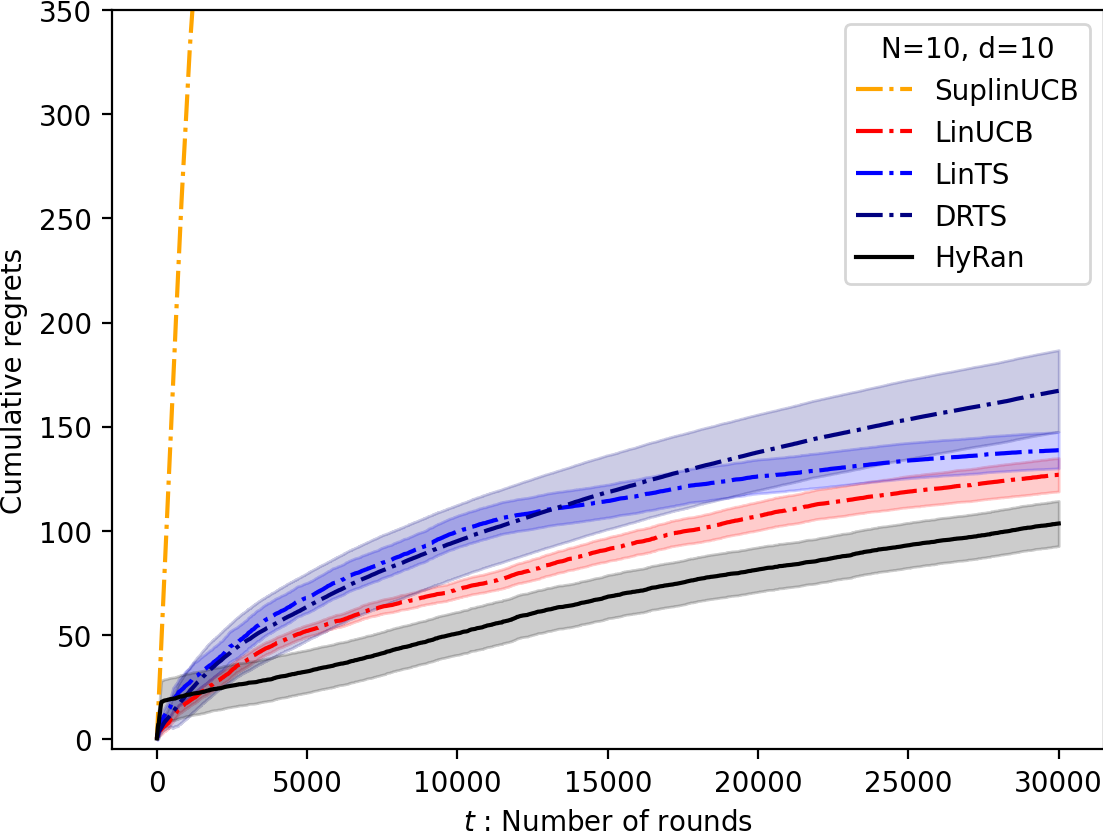}
\includegraphics[width=0.32\linewidth]{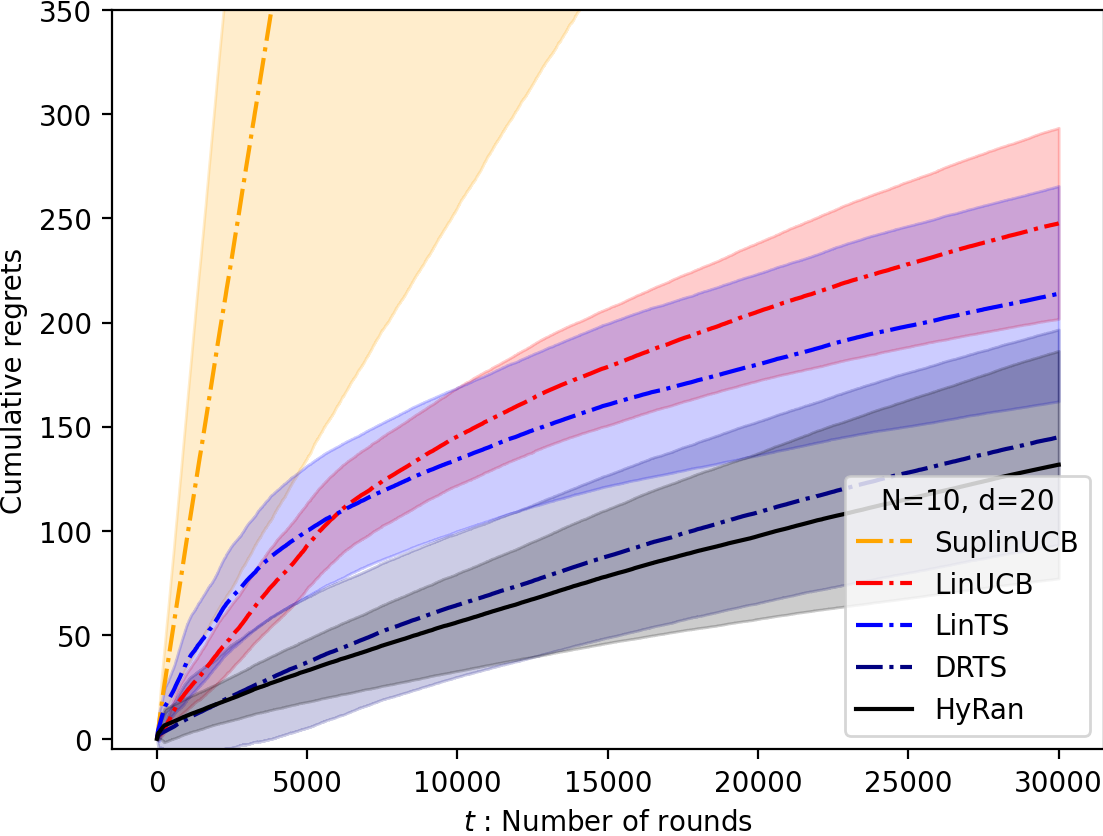}
\\
 \includegraphics[width=0.32\linewidth]{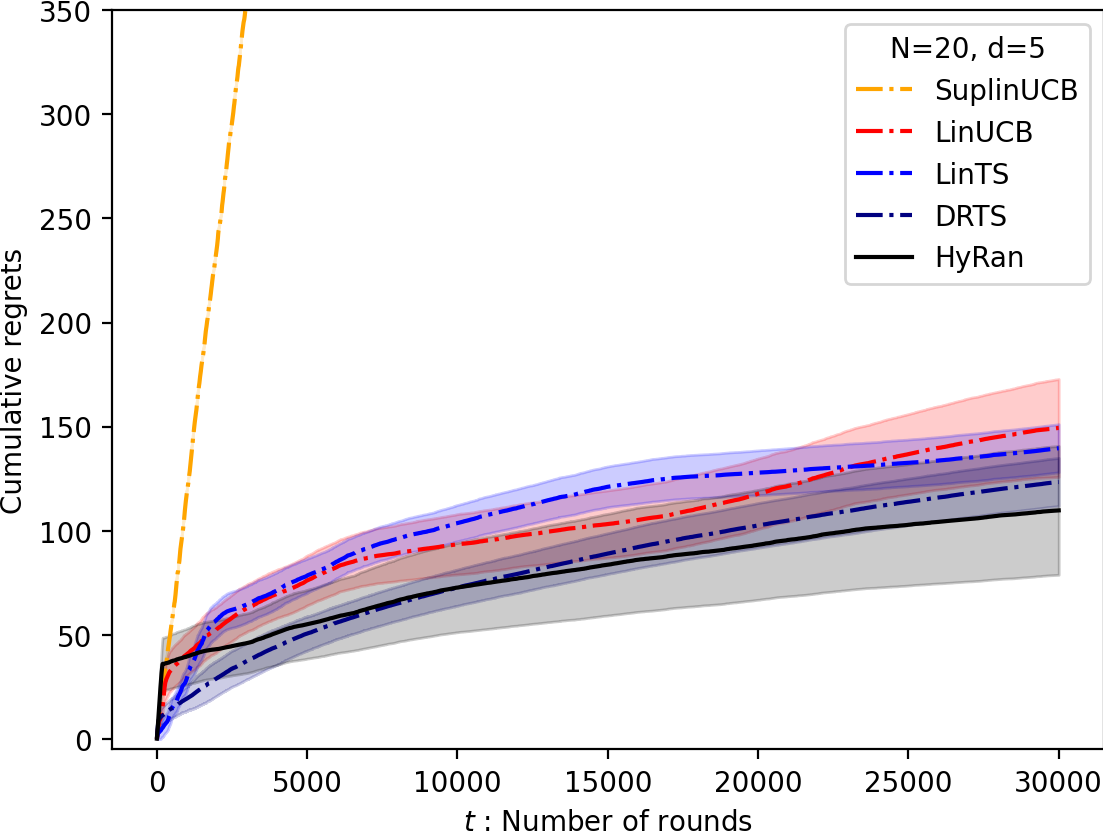}
\includegraphics[width=0.32\linewidth]{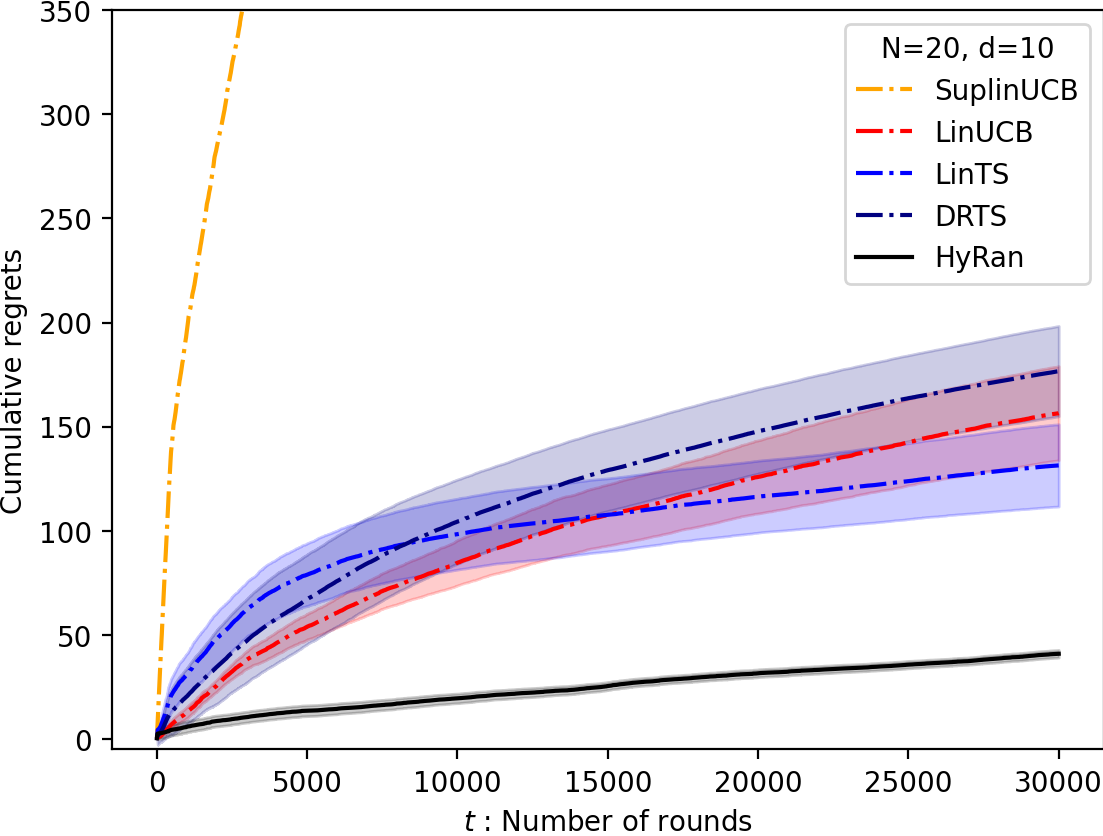}
\includegraphics[width=0.32\linewidth]{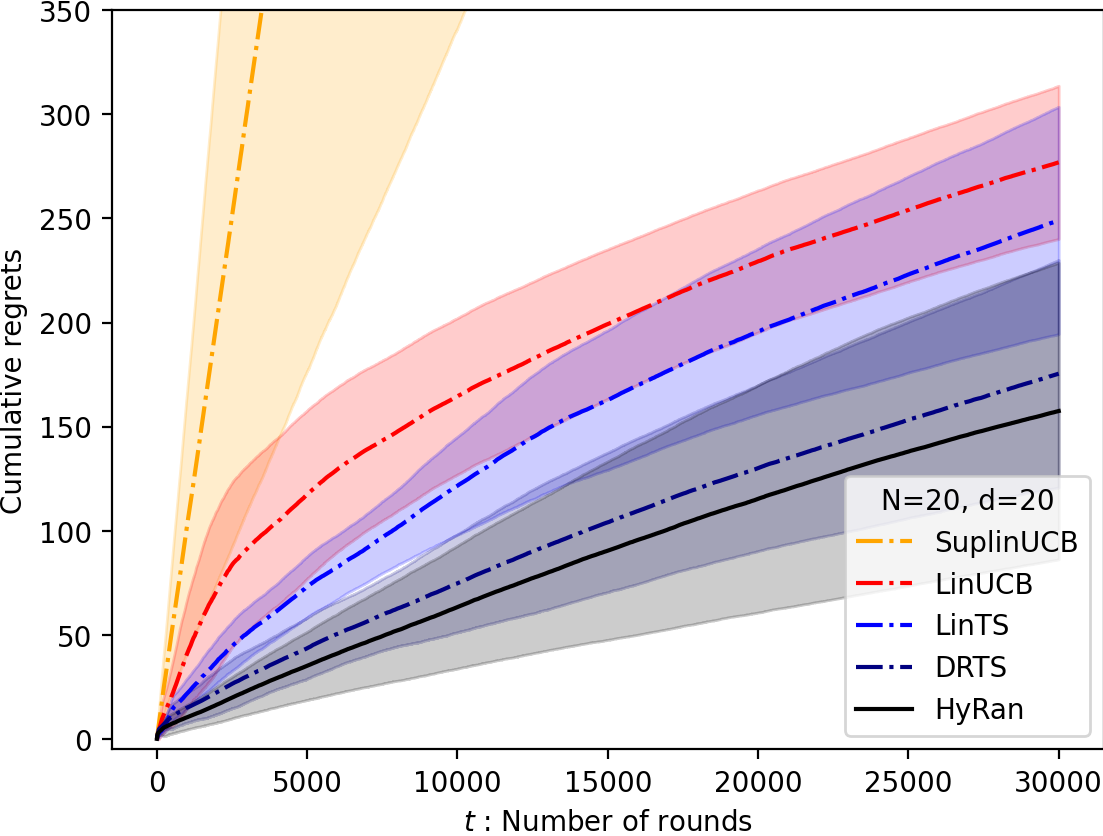}
\caption{ A comparison of cumulative regrets of \texttt{SupLinUCB}, \texttt{LinTS},
\texttt{LinUCB, DRTS} and \texttt{HyRan Bandit}. Each curve shows
the cumulative regret as a function of rounds, averaged over 20 repeated
experiments. The scale of $y$-axis is set to be equivalent in each
row for the comparison of the regret as $d$ increases. The standard
deviations of \texttt{SupLinUCB} in $d=5$ and $d=10$ are too large
to present and omitted. }
\label{fig:results} 
\end{figure*}

Theorem~\ref{thm:Estimation_error} is a self-normalized bound for
the \texttt{HyRan} estimator, which is a crucial element in our regret
analysis. Compared to the widely-used self-normalization bound (Theorem~2
in \citet{abbasi2011improved}) in the contextual bandit literature,
the estimation error bound (\ref{eq:estimation_error_bound}) is self-normalized
by the covariance matrix constructed by the contexts of all arms,
not just selected contexts. The self-normalized bound is derived by
using the pairs of pseudo reward $\Tildereward i{\tau}$ defined in
\eqref{eq:pseudo_reward} and contexts $\Context i{\tau}$ for all
arms $i\in[N]$ and $\tau\in\Psi_{t}$, instead of using just the
pairs of selected arms. The full usage of pseudo rewards and contexts
enables us to take advantage of the new decomposition of the regret
in \eqref{eq:regret_decomposition}, which derives a $O(\sqrt{dT\log T})$
regret bound.

The last concern regarding our regret bound is the size of $\Psi_{t}$.
To obtain a regret bound sublinear to $T$, we need to make sure that
the sum of the subsampled rounds satisfies $\sum_{t=1}^{T}\abs{\Psi_{t}}^{-1/2}=O(\sqrt{T})$.
In the following Lemma, we show this by proving that the size of the
selected subset $\Psi_{t}$ is $\Omega(t)$ with high probability. 
\begin{lem}
\label{lem:psi_size} Let $\Psi_{t}$ be a subset of $[t]$ determined
by the Algorithm \ref{alg:HyRan} at round $t$. For any $\epsilon\in(0,1)$,
with probability at least $1-\delta$, 
\begin{equation}
\abs{\Psi_{t}}\ge\epsilon pt,\label{eq:psi_size}
\end{equation}
for all $t\ge\frac{2}{p\left(1-\epsilon\right)^{2}}\log\frac{T}{\delta}$. 
\end{lem}

With~\eqref{eq:psi_size}, we guarantee the rate of the regret bound
is sub-linear with respect to the total round $T$.

\subsection{Matching Lower Bound}

\label{subsec:lower_bound}

Regarding the lower bounds of the linear contextual bandit, a $\Omega(d\sqrt{T})$
bound has been proven for linear bandits with infinitely many arms
\citep{dani2008stochastic,rusmevichientong2010linearly,lattimore2020bandit}.
When the number of arms is finite, the derived lower bound of the
cumulative regret is $\Omega(\sqrt{dT})$ \citep{chu2011contextual}.
Recently in \citet{li2019nearly}, a lower bound $\Omega(\sqrt{dT\log T\log N})$
was shown when $N\le2^{d/2}$. These lower bounds are derived by finding
the settings of contexts and parameters that make the algorithm difficult
to reduce the regret. However, the problem settings of the existing
lower bounds do not satisfy Assumptions \ref{assum:iid_context} in
our problem setting. In the following theorem, we prove a lower bound
which is valid under Assumptions \ref{assum:boundedness}-\ref{assum:iid_context}.
\begin{thm}
\label{thm:lower_bound} Assume $2\le d\le N<\infty$ and $T\ge d/4$.
Then there exists a distribution of contexts, $\mathcal{P}_{\mathcal{X}}$,
a distribution of noise, $\Error it$ and $\beta^{*}$, which satisfies
Assumptions \ref{assum:boundedness}-\ref{assum:iid_context} and
for any bandit algorithms that selects $\Action t$, 
\begin{equation}
\Expectation_{\beta^{*}}R(T)\ge\frac{1}{8}\sqrt{dT}.\label{eq:lower_bound}
\end{equation}
\end{thm}

We prove that the rate of $\Omega(\sqrt{dT})$ cannot be improved
even under the stochastic assumptions on contexts (e.g., Assumption~\ref{assum:iid_context}).
The lower bound in Theorem~\ref{thm:lower_bound} matches with our
regret upper bound for \texttt{HyRan Bandit} established in Theorem~\ref{thm:regret_bound}
up to the logarithmic factor. Therefore, our proposed algorithm \texttt{HyRan
Bandit} is provably near-optimal, i.e., optimal up to the logarithmic
factor. To our knowledge, all of the existing near-optimal linear
contextual bandit algorithms are based on the framework of \citet{auer2002using}
(e.g., \texttt{SupLinUCB} and \texttt{VCL-SupLinUCB}). Our proposed
algorithm is the first algorithm that achieves near-optimality without
relying on this existing framework.

Despite the lower bound is derived under Assumption \ref{assum:iid_context}
related to the factor $\phi>0$ , our lower bound \eqref{eq:lower_bound}
does not have $\phi$. This is because the lower bound depends only
on the number of orthogonal vectors in the contexts space $\Real^{d}$,
not the value of $\phi>0$. 


\section{NUMERICAL EXPERIMENTS}

\label{sec:experiments}

In this section, we compare the performances of the five linear contextual
bandit algorithms: \texttt{SupLinUCB} \citep{chu2011contextual},
\texttt{LinUCB} \citep{li2010contextual}, \texttt{LinTS} \citep{agrawal2013thompson},
\texttt{DRTS} \citep{kim2021doubly} and our proposed method, \texttt{HyRan
Bandit}. For simulation, the number of arms $N$ is set to $10$ or
$20$, and the dimension of contexts $d$ is set to $5$, $10$ and
$20$. Let $X_{i,t}^{(1)},\ldots,X_{i,t}^{(d)}$ be the $d$ elements
of a context $\Context it$. For $j=1,\ldots,d-1$, we independently
generate $(X_{1,t}^{(j)},\cdots,X_{N,t}^{(j)})$ from a normal distribution
$\mathcal{N}(\mu_{N},V_{N})$ with mean $\mu_{10}\!=\!(-10,-8,\cdots,-2,2,\cdots8,-10)^{T}$,
or $\mu_{20}=(-20,-18,\cdots,-2,2,\cdots,18,20)^{T}$. To impose correlation
among each arms the covariance matrix $V_{N}\in\mathbb{R}^{N\times N}$
is set as $V(i,i)=1$ for every $i$ and $V(i,k)=0.5$ for every $i\neq k$.
Then, for each arm $i\in[N]$, we randomly select a generated element
$X_{i,t}^{(j)}$ and append it to the last element, i.e. $X_{i,t}^{(d)}$
is the same as one of $X_{i,t}^{(1)},\ldots,X_{i,t}^{(d-1)}$. This
setting is to impose a severe multicollinearity on each contexts.
Finally, we truncated the sampled contexts to satisfy $\|\Context it\|_{2}\le1$.
To generate the stochastic rewards, we sample $\Error it$ independently
from $\mathcal{N}(0,1)$. Each element of $\beta^{*}$ is sampled
from a uniform distribution, $\mathcal{U}(-{1}/{\sqrt{d}},{1}/{\sqrt{d}})$
at the beginning of each instance and stays fixed during experiments.
About the set of hyperparameters, \texttt{LinTS}, \texttt{LinUCB}, \texttt{SupLinUCB} and \texttt{DRTS} searches $\alpha$ (or $v$) in $\{0.001,0.01,0.1,1\}$.
In \texttt{HyRan Bandit} we set $\lambda_{t}:=d\log(t+1)^{2}$ to
be consistent with the theoretical results and $p$ to be in \{0.5,0.65,0.8,0.95\}.
We optimize the hyperparameters over the grid set and report the best performance for each algorithm. 
Figure~\ref{fig:results} shows the average of the cumulative regrets over the horizon length $T=30000$ with 20 repeated experiments. 
The experimental results demonstrate that \texttt{HyRan Bandit} performs better than the benchmarks in all of the cases and shows superior performances as the context dimension increases. 
The worst performance of \texttt{SupLinUCB} is mainly because its estimator does not include rewards in exploitation rounds. 


\section{CONCLUSION}

We address a long-standing research question of whether a practical
algorithm can achieve near-optimality for linear contextual bandits.
We show that our proposed algorithm achieves $\tilde{O}(\sqrt{dT})$
regret upper bound which matches the lower bound under our problem
setting. We empirically evaluate our algorithm to support our theoretical
claims and show that the practical performance of our algorithm outperforms the existing methods, hence achieving both provable near-optimality and practicality.

\subsubsection*{Acknowledgements}
This work is supported by the National Research Foundation of Korea (NRF) grant funded by the Korea government (MSIT, No.2020R1A2C1A01011950) (WK and MCP) and (MSIT, No. 2022R1C1C1006859) (MO), and also supported by Naver (MO) and Hyundai Chung Mong-koo scholarship foundation (WK).

\bibliographystyle{plainnat}
\bibliography{ref}

\begin{thebibliography}{32}
\providecommand{\natexlab}[1]{#1}
\providecommand{\url}[1]{\texttt{#1}}
\expandafter\ifx\csname urlstyle\endcsname\relax
  \providecommand{\doi}[1]{doi: #1}\else
  \providecommand{\doi}{doi: \begingroup \urlstyle{rm}\Url}\fi

\bibitem[Abbasi-Yadkori et~al.(2011)Abbasi-Yadkori, P{\'a}l, and
  Szepesv{\'a}ri]{abbasi2011improved}
Yasin Abbasi-Yadkori, D{\'a}vid P{\'a}l, and Csaba Szepesv{\'a}ri.
\newblock Improved algorithms for linear stochastic bandits.
\newblock In \emph{Advances in Neural Information Processing Systems}, pages
  2312--2320, 2011.

\bibitem[Abe and Long(1999)]{abe1999associative}
Naoki Abe and Philip~M Long.
\newblock Associative reinforcement learning using linear probabilistic
  concepts.
\newblock In \emph{ICML}, 1999.

\bibitem[Abeille et~al.(2017)Abeille, Lazaric, et~al.]{abeille2017linear}
Marc Abeille, Alessandro Lazaric, et~al.
\newblock Linear thompson sampling revisited.
\newblock \emph{Electronic Journal of Statistics}, 11\penalty0 (2):\penalty0
  5165--5197, 2017.

\bibitem[Agrawal and Goyal(2013)]{agrawal2013thompson}
Shipra Agrawal and Navin Goyal.
\newblock Thompson sampling for contextual bandits with linear payoffs.
\newblock In \emph{International Conference on Machine Learning}, pages
  127--135, 2013.

\bibitem[Amani et~al.(2019)Amani, Alizadeh, and Thrampoulidis]{amani2019linear}
Sanae Amani, Mahnoosh Alizadeh, and Christos Thrampoulidis.
\newblock Linear stochastic bandits under safety constraints.
\newblock In \emph{Advances in Neural Information Processing Systems}, pages
  9252--9262, 2019.

\bibitem[Auer(2002a)]{auer2002using}
Peter Auer.
\newblock Using confidence bounds for exploitation-exploration trade-offs.
\newblock \emph{Journal of Machine Learning Research}, 3\penalty0
  (Nov):\penalty0 397--422, 2002a.

\bibitem[Auer et~al.(2002b)Auer, Cesa-Bianchi, Freund, and
  Schapire]{auer2002nonstochastic}
Peter Auer, Nicolo Cesa-Bianchi, Yoav Freund, and Robert~E Schapire.
\newblock The nonstochastic multiarmed bandit problem.
\newblock \emph{SIAM journal on computing}, 32\penalty0 (1):\penalty0 48--77,
  2002b.

\bibitem[Bang and Robins(2005)]{bang2005doubly}
Heejung Bang and James~M Robins.
\newblock Doubly robust estimation in missing data and causal inference models.
\newblock \emph{Biometrics}, 61\penalty0 (4):\penalty0 962--973, 2005.

\bibitem[Bastani and Bayati(2020)]{bastani2020online}
Hamsa Bastani and Mohsen Bayati.
\newblock Online decision making with high-dimensional covariates.
\newblock \emph{Operations Research}, 68\penalty0 (1):\penalty0 276--294, 2020.

\bibitem[Bastani et~al.(2021)Bastani, Bayati, and Khosravi]{bastani2021mostly}
Hamsa Bastani, Mohsen Bayati, and Khashayar Khosravi.
\newblock Mostly exploration-free algorithms for contextual bandits.
\newblock \emph{Management Science}, 67\penalty0 (3):\penalty0 1329--1349,
  2021.

\bibitem[B{\"u}hlmann and Van De~Geer(2011)]{buhlmann2011statistics}
Peter B{\"u}hlmann and Sara Van De~Geer.
\newblock \emph{Statistics for high-dimensional data: methods, theory and
  applications}.
\newblock Springer Science \& Business Media, 2011.

\bibitem[Chu et~al.(2011)Chu, Li, Reyzin, and Schapire]{chu2011contextual}
Wei Chu, Lihong Li, Lev Reyzin, and Robert Schapire.
\newblock Contextual bandits with linear payoff functions.
\newblock In \emph{Proceedings of the Fourteenth International Conference on
  Artificial Intelligence and Statistics}, pages 208--214, 2011.

\bibitem[Dani et~al.(2008)Dani, Hayes, and Kakade]{dani2008stochastic}
Varsha Dani, Thomas Hayes, and Sham Kakade.
\newblock Stochastic linear optimization under bandit feedback.
\newblock In \emph{21st Annual Conference on Learning Theory}, pages 355--366,
  01 2008.

\bibitem[Dimakopoulou et~al.(2019)Dimakopoulou, Zhou, Athey, and
  Imbens]{dimakopoulou2019balanced}
Maria Dimakopoulou, Zhengyuan Zhou, Susan Athey, and Guido Imbens.
\newblock Balanced linear contextual bandits.
\newblock In \emph{Proceedings of the AAAI Conference on Artificial
  Intelligence}, volume~33, pages 3445--3453, 2019.

\bibitem[Goldenshluger and Zeevi(2013)]{goldenshluger2013linear}
Alexander Goldenshluger and Assaf Zeevi.
\newblock A linear response bandit problem.
\newblock \emph{Stochastic Systems}, 3\penalty0 (1):\penalty0 230--261, 2013.

\bibitem[Kim and Paik(2019)]{kim2019doubly}
Gisoo Kim and Myunghee~Cho Paik.
\newblock Doubly-robust lasso bandit.
\newblock In \emph{Advances in Neural Information Processing Systems}, pages
  5869--5879, 2019.

\bibitem[Kim et~al.(2021)Kim, Kim, and Paik]{kim2021doubly}
Wonyoung Kim, Gi-Soo Kim, and Myunghee~Cho Paik.
\newblock Doubly robust thompson sampling with linear payoffs.
\newblock In A.~Beygelzimer, Y.~Dauphin, P.~Liang, and J.~Wortman Vaughan,
  editors, \emph{Advances in Neural Information Processing Systems}, 2021.

\bibitem[Kim et~al.(2022)Kim, Lee, and Paik]{kim2022double}
Wonyoung Kim, Kyungbok Lee, and Myunghee~Cho Paik.
\newblock Double doubly robust thompson sampling for generalized linear
  contextual bandits.
\newblock \emph{arXiv preprint arXiv:2209.06983}, 2022.

\bibitem[Kontorovich and Ramanan(2008)]{kontorovich2008concentration}
Leonid~Aryeh Kontorovich and Kavita Ramanan.
\newblock Concentration inequalities for dependent random variables via the
  martingale method.
\newblock \emph{The Annals of Probability}, 36\penalty0 (6):\penalty0
  2126--2158, 2008.

\bibitem[Kveton et~al.(2020)Kveton, Szepesv{\'{a}}ri, Ghavamzadeh, and
  Boutilier]{perturb20akveton}
Branislav Kveton, Csaba Szepesv{\'{a}}ri, Mohammad Ghavamzadeh, and Craig
  Boutilier.
\newblock Perturbed-history exploration in stochastic linear bandits.
\newblock In Ryan~P. Adams and Vibhav Gogate, editors, \emph{Proceedings of The
  35th Uncertainty in Artificial Intelligence Conference}, volume 115 of
  \emph{Proceedings of Machine Learning Research}, pages 530--540. PMLR, 22--25
  Jul 2020.

\bibitem[Lai and Robbins(1985)]{lai1985asymptotically}
Tze~Leung Lai and Herbert Robbins.
\newblock Asymptotically efficient adaptive allocation rules.
\newblock \emph{Advances in applied mathematics}, 6\penalty0 (1):\penalty0
  4--22, 1985.

\bibitem[Lattimore and Szepesv{\'a}ri(2020)]{lattimore2020bandit}
Tor Lattimore and Csaba Szepesv{\'a}ri.
\newblock \emph{Bandit algorithms}.
\newblock Cambridge University Press, 2020.

\bibitem[Lee et~al.(2016)Lee, Peres, and Smart]{lee2016}
James~R. Lee, Yuval Peres, and Charles~K. Smart.
\newblock A gaussian upper bound for martingale small-ball probabilities.
\newblock \emph{Ann. Probab.}, 44\penalty0 (6):\penalty0 4184--4197, 11 2016.
\newblock \doi{10.1214/15-AOP1073}.

\bibitem[Li et~al.(2010)Li, Chu, Langford, and Schapire]{li2010contextual}
Lihong Li, Wei Chu, John Langford, and Robert~E Schapire.
\newblock A contextual-bandit approach to personalized news article
  recommendation.
\newblock In \emph{Proceedings of the 19th international conference on World
  wide web}, pages 661--670, 2010.

\bibitem[Li et~al.(2017)Li, Lu, and Zhou]{li2017provably}
Lihong Li, Yu~Lu, and Dengyong Zhou.
\newblock Provably optimal algorithms for generalized linear contextual
  bandits.
\newblock In \emph{Proceedings of the 34th International Conference on Machine
  Learning-Volume 70}, pages 2071--2080. JMLR. org, 2017.

\bibitem[Li et~al.(2019)Li, Wang, and Zhou]{li2019nearly}
Yingkai Li, Yining Wang, and Yuan Zhou.
\newblock Nearly minimax-optimal regret for linearly parameterized bandits.
\newblock In \emph{Conference on Learning Theory}, pages 2173--2174. PMLR,
  2019.

\bibitem[Oh et~al.(2021)Oh, Iyengar, and Zeevi]{oh2021sparsity}
Min-hwan Oh, Garud Iyengar, and Assaf Zeevi.
\newblock Sparsity-agnostic lasso bandit.
\newblock In \emph{International Conference on Machine Learning}, pages
  8271--8280. PMLR, 2021.

\bibitem[Robins et~al.(1994)Robins, Rotnitzky, and Zhao]{robins1994}
James~M. Robins, Andrea Rotnitzky, and Lue~Ping Zhao.
\newblock Estimation of regression coefficients when some regressors are not
  always observed.
\newblock \emph{Journal of the American Statistical Association}, 89\penalty0
  (427):\penalty0 846--866, 1994.
\newblock ISSN 01621459.

\bibitem[Rusmevichientong and Tsitsiklis(2010)]{rusmevichientong2010linearly}
Paat Rusmevichientong and John~N Tsitsiklis.
\newblock Linearly parameterized bandits.
\newblock \emph{Mathematics of Operations Research}, 35\penalty0 (2):\penalty0
  395--411, 2010.

\bibitem[Thompson(1933)]{1933Thompson}
William~R. Thompson.
\newblock On the likelihood that one unknown probability exceeds another in
  view of the evidence of two samples.
\newblock \emph{Biometrika}, 25\penalty0 (3/4):\penalty0 285--294, 1933.
\newblock ISSN 00063444.

\bibitem[Valko et~al.(2014)Valko, Munos, Kveton, and
  Koc{\'a}k]{valko2014spectral}
Michal Valko, R{\'e}mi Munos, Branislav Kveton, and Tom{\'a}{\v{s}} Koc{\'a}k.
\newblock Spectral bandits for smooth graph functions.
\newblock In \emph{International Conference on Machine Learning}, pages 46--54.
  PMLR, 2014.

\bibitem[van~der Vaart and Wellner(1996)]{vanderVaart1996}
Aad~W. van~der Vaart and Jon~A. Wellner.
\newblock \emph{Symmetrization and Measurability}, pages 107--121.
\newblock Springer New York, New York, NY, 1996.
\newblock ISBN 978-1-4757-2545-2.
\newblock \doi{10.1007/978-1-4757-2545-2_15}.

\end{thebibliography}

\clearpage{}

\onecolumn

\appendix

\section{MISSING PROOFS}

\subsection{Technical lemmas}
\begin{lem}
\label{lem:hilbert_to_R2} \citet[Lemma 2.3]{lee2016} Let $\left\{ N_{t}\right\} $
be a martingale on a Hilbert space $(\mathcal{H},\norm{\cdot}_{\mathcal{H}})$.
Then there exists a $\Real^{2}$-valued martingale $\left\{ M_{t}\right\} $
such that for any time $t\ge0$, $\norm{M_{t}}_{2}=\norm{N_{t}}_{\mathcal{H}}$
and $\norm{M_{t+1}-M_{t}}_{2}=\norm{N_{t+1}-N_{t}}_{\mathcal{H}}$. 
\end{lem}

\begin{lem}
\label{lem:azuma} (Azuma-Hoeffding) If a super-martingale $(Y_{t};t\ge0)$
corresponding to filtration $\Filtration t$, satisfies $\abs{Y_{t}-Y_{t-1}}\le c_{t}$
for some constant $c_{t}$, for all $t=1,\ldots,T$, then for any
$a\ge0$, 
\[
\Probability\left(Y_{T}-Y_{0}\ge a\right)\le e^{-\frac{a^{2}}{2\sum_{t=1}^{T}c_{t}^{2}}}.
\]
\end{lem}

\subsection{Proof of Theorem \ref{thm:regret_bound}}
\begin{proof}
{[}Step 1. Regret decomposition{]} For each $t\in[T]$, define the
event 
\[
\begin{split}A_{t} & :=\left\{ \abs{\Psi_{t}}>\frac{1}{2}pt\right\} ,\\
B_{t} & :=\left\{ \norm{\Estimator t-\beta^{*}}_{V_{t}}\le\sqrt{\lambda_{t}}+\left(\frac{4\sqrt{2}}{1-p}+\frac{\sigma}{p}\right)\sqrt{d\log\frac{4t^{2}}{\delta}}\right\} ,\\
C_{t} & :=\left\{ \norm{\Estimator t-\beta^{*}}_{2}\le1+\frac{4\sqrt{2}}{1-p}+\frac{\sigma}{p}:=D_{p,\sigma}\right\} .
\end{split}
\]
The three events have an explicit relationship as follows: In the
proof of Theorem \ref{thm:Estimation_error}, Lemma \ref{lem:psi_size}
and Lemma \ref{lem:imputation_estimator_bound}, the event $B_{t}$
requires $A_{t}$, i.e. $A_{t}\subseteq B_{t}$. Under the event $B_{t}$,
setting $\lambda_{t}=d\log\frac{4t^{2}}{\delta}$ gives 
\begin{align*}
\norm{\Estimator t-\beta^{*}}_{2}\le & \sqrt{\left(\Estimator t-\beta^{*}\right)^{T}V_{t}^{\frac{1}{2}}V_{t}^{-1}V_{t}^{\frac{1}{2}}\left(\Estimator t-\beta^{*}\right)}\\
\le & \sqrt{\Maxeigen{V_{t}^{-1}}}\norm{\Estimator t-\beta^{*}}_{V_{t}}\\
\le & \lambda_{t}^{-\frac{1}{2}}\left(\sqrt{\lambda_{t}}+\left(\frac{4\sqrt{2}}{1-p}+\frac{\sigma}{p}\right)\sqrt{d\log\frac{4t^{2}}{\delta}}\right)\\
\le & D_{p,\sigma},
\end{align*}
which implies $C_{t}$. Set $\mathcal{E}:=\max\left\{ \frac{8}{p}\log\frac{T}{\delta},C_{p,\sigma}N^{2}\phi^{-4}\log\frac{2T}{\delta}\right\} $,
where $C_{p,\sigma}$ is defined in \eqref{eq:C_p_sigma}. By Theorem
\ref{thm:Estimation_error} we have 
\begin{equation}
\Probability\left(\bigcap_{t\ge\mathcal{E}}\left\{ A_{t}\cap B_{t}\cap C_{t}\right\} \right)\ge1-6\delta.\label{eq:abc_event}
\end{equation}
By Lemma \ref{lem:regret_decomposition}, for each $t\ge\mathcal{E}$,
\begin{align*}
\Regret t & \le2\left\{ \Maxresidual{\SetofContexts t}{\Estimator{t-1}}-\CE{\Maxresidual{\SetofContexts t}{\Estimator{t-1}}}{\mathcal{G}_{t-1}}\right\} \\
 & +2\left\{ \CE{\Maxresidual{\SetofContexts t}{\Estimator{t-1}}}{\mathcal{G}_{t-1}}-\frac{1}{\abs{\Psi_{t-1}}}\sum_{\tau\in\Psi_{t-1}}\Maxresidual{\SetofContexts{\tau}}{\Estimator{t-1}}\right\} \\
 & +\frac{2}{\sqrt{\abs{\Psi_{t-1}}}}\norm{\beta^{*}-\Estimator{t-1}}_{V_{t-1}}.
\end{align*}
Let 
\begin{equation}
\begin{split}R_{1}(t):= & 2\left\{ \Maxresidual{\SetofContexts t}{\Estimator{t-1}}-\CE{\Maxresidual{\SetofContexts t}{\Estimator{t-1}}}{\mathcal{G}_{t-1}}\right\} ,\\
R_{2}(t):= & 2\left\{ \CE{\Maxresidual{\SetofContexts t}{\Estimator{t-1}}}{\mathcal{G}_{t-1}}-\frac{1}{\abs{\Psi_{t-1}}}\sum_{\tau\in\Psi_{t-1}}\Maxresidual{\SetofContexts{\tau}}{\Estimator{t-1}}\right\} ,\\
R_{3}(t):= & \frac{2}{\sqrt{\abs{\Psi_{t-1}}}}\norm{\beta^{*}-\Estimator{t-1}}_{V_{t-1}}.
\end{split}
\label{eq:R_decomposed_terms}
\end{equation}

{[}Step 2. Bounding $R_{1}(t)${]} Let us bound $R_{1}(t)$. Since
the event $C_{t}$ is $\mathcal{G}_{t}$-measurable for each $t\in[T]$,
we have 
\[
R_{1}(t)\Indicator{C_{t-1}}=2\left\{ \Maxresidual{\SetofContexts t}{\Estimator{t-1}}\Indicator{C_{t-1}}-\CE{\Maxresidual{\SetofContexts t}{\Estimator{t-1}}\Indicator{C_{t-1}}}{\mathcal{G}_{t-1}}\right\} .
\]
By Assumption 1, 
\begin{align*}
\Maxresidual{\SetofContexts t}{\Estimator{t-1}}\Indicator{C_{t-1}}:= & \max_{i\in[N]}\abs{\Context it^{T}\left(\Estimator{t-1}-\beta^{*}\right)}\Indicator{C_{t-1}}\\
\le & \max_{i\in[N]}\norm{\Context it}_{2}\norm{\Estimator{t-1}-\beta^{*}}_{2}\Indicator{C_{t-1}}\\
\le & \norm{\Estimator{t-1}-\beta^{*}}_{2}\Indicator{C_{t-1}}\\
\le & D_{p,\sigma}.
\end{align*}
Thus, $\abs{R_{1}(t)\Indicator{C_{t-1}}}\le4D_{p,\sigma}$. Since
$R_{1}(t)\Indicator{C_{t-1}}$ is $\mathcal{G}_{t}$-measurable and
\[
\CE{R_{1}(t)\Indicator{C_{t-1}}}{\mathcal{G}_{t-1}}=0,
\]
we can use Lemma \ref{lem:azuma} to have 
\begin{equation}
\sum_{t>\mathcal{E}}R_{1}(t)\Indicator{C_{t-1}}\le4D_{p,\sigma}\sqrt{2T\log\frac{1}{\delta}},\label{eq:R_1_bound}
\end{equation}
with probability at least $1-\delta$.

{[}Step 3. Bounding $R_{2}(t)${]} Now we bound $R_{2}(t)$. By Lemma
\ref{lem:expectation_bound} with probability at least $1-\delta/T$,
\begin{align*}
R_{2}(t)\Indicator{A_{t-1}\cap C_{t-1}} & \le2\Indicator{A_{t-1}}\!\sup_{\norm{\beta_{1}-\beta^{*}}_{2}\le D_{p,\sigma}}\!\abs{\CE{\Maxresidual{\SetofContexts t}{\beta_{1}}}{\mathcal{G}_{t-1}}-\frac{1}{\abs{\Psi_{t-1}}}\sum_{\tau\in\Psi_{t-1}}\Maxresidual{\SetofContexts{\tau}}{\beta_{1}}}\\
 & \le\left(\frac{3\delta D_{p,\sigma}}{T}+8D_{p,\sigma}\sqrt{\frac{1}{\abs{\Psi_{t-1}}}}\sqrt{d\log\frac{2T}{\delta}}\right)\Indicator{A_{t-1}}\\
 & \le\frac{3\delta D_{p,\sigma}}{T}+8D_{p,\sigma}\sqrt{\frac{2}{pt}}\sqrt{d\log\frac{2T}{\delta}}.
\end{align*}
Thus, with probability at least $1-\delta$, 
\begin{equation}
\sum_{t>\mathcal{E}}R_{2}(t)\Indicator{A_{t-1}\cap C_{t-1}}\le3\delta D_{p,\sigma}+\frac{16\sqrt{2}D_{p,\sigma}}{\sqrt{p}}\sqrt{dT\log\frac{2T}{\delta}}.\label{eq:R_2_bound}
\end{equation}

{[}Step 4. Bounding $R_{3}(t)${]} To bound $R_{3}(t)$, 
\begin{align*}
R_{3}(t)\Indicator{A_{t-1}\cap B_{t-1}}\le & \frac{2\sqrt{2}}{\sqrt{pt}}\left(1+\frac{4C}{1-p}+\frac{\sigma}{p}\right)\sqrt{d\log\frac{4t^{2}}{\delta}}\\
= & \frac{2\sqrt{2}}{\sqrt{pt}}D_{p,\sigma}\sqrt{d\log\frac{4t^{2}}{\delta}}.
\end{align*}
and 
\begin{equation}
\sum_{t>\mathcal{E}}R_{3}(t)\Indicator{A_{t-1}\cap B_{t-1}}\le\frac{8D_{p,\sigma}}{\sqrt{p}}\sqrt{dT\log\frac{2T}{\delta}},\label{eq:R_3_bound}
\end{equation}
holds almost surely. 

{[}Step 5. Collecting the bounds{]} For any $x>2\mathcal{E}$, 
\begin{align*}
\Probability\left(R(T)>x\right) & \le\Probability\left(2\mathcal{E}+\sum_{t>\mathcal{E}}\Regret t>x\right)\\
 & =\Probability\left(2\mathcal{E}+\sum_{t>\mathcal{E}}R_{1}(t)+R_{2}(t)+R_{3}(t)>x\right)\\
 & \le\Probability\!\left(2\mathcal{E}\!+\!\sum_{t>\mathcal{E}}R_{1}(t)\Indicator{C_{t-1}}\!+\!R_{2}(t)\Indicator{A_{t-1}\!\cap\!C_{t-1}}\!+\!R_{3}(t)\Indicator{A_{t-1}\!\cap\!B_{t-1}}>x\right)\\
 & \quad+\Probability\left(\bigcup_{t\ge\mathcal{E}}\left\{ A_{t}^{c}\cup B_{t}^{c}\cup C_{t}^{c}\right\} \right)\\
 & \le\Probability\!\left(2\mathcal{E}\!+\!\sum_{t>\mathcal{E}}R_{1}(t)\Indicator{C_{t-1}}\!+\!R_{2}(t)\Indicator{A_{t-1}\!\cap\!C_{t-1}}\!+\!R_{3}(t)\Indicator{A_{t-1}\!\cap\!B_{t-1}}>x\right)\\
 & \quad+6\delta,
\end{align*}
where the last inequality holds due to \eqref{eq:abc_event}. Setting
\[
x=2\mathcal{E}+4D_{p,\sigma}\sqrt{2T\log\frac{1}{\delta}}+3\delta D_{p,\sigma}+\frac{16\sqrt{2}D_{p,\sigma}}{\sqrt{p}}\sqrt{dT\log\frac{2T}{\delta}}+\frac{8D_{p,\sigma}}{\sqrt{p}}\sqrt{dT\log\frac{2T}{\delta}},
\]
gives 
\begin{align*}
\Probability\left(R(T)>x\right)\le & 6\delta+\Probability\left(\sum_{t>\mathcal{E}}R_{1}(t)\Indicator{C_{t-1}}>4D_{p,\sigma}\sqrt{2T\log\frac{1}{\delta}}\right)\\
 & +\Probability\left(\sum_{t>\mathcal{E}}R_{2}(t)\Indicator{A_{t-1}\cap C_{t-1}}>3\delta D_{p,\sigma}+\frac{16\sqrt{2}D_{p,\sigma}}{\sqrt{p}}\sqrt{dT\log\frac{2T}{\delta}}\right)\\
 & +\Probability\left(\sum_{t>\mathcal{E}}R_{3}(t)\Indicator{A_{t-1}\cap C_{t-1}}>\frac{8D_{p,\sigma}}{\sqrt{p}}\sqrt{dT\log\frac{2T}{\delta}}\right)\\
\le & 8\delta,
\end{align*}
where the inequality holds due to \eqref{eq:R_decomposed_terms}-\eqref{eq:R_3_bound}. 
\end{proof}

\subsection{Proof of Lemma \ref{lem:regret_decomposition}}
\begin{proof}
By the definition of $\Action t$, we have 
\[
\begin{split}\Regret{t+1}= & \left(\Context{\Optimalarm{t+1}}{t+1}-\Context{\Action{t+1}}{t+1}\right)^{T}\left(\beta^{*}-\Estimator t\right)+\left(\Context{\Optimalarm{t+1}}{t+1}-\Context{\Action{t+1}}{t+1}\right)^{T}\Estimator t\\
\le & \left(\Context{\Optimalarm{t+1}}{t+1}-\Context{\Action{t+1}}{t+1}\right)^{T}\left(\beta^{*}-\Estimator t\right)\\
\le & 2\max_{i\in[N]}\abs{\Context i{t+1}^{T}\left(\Estimator t-\beta^{*}\right)},
\end{split}
\]
which gives $\Regret{t+1}\le2\Maxresidual{\SetofContexts{t+1}}{\Estimator t}$.
Adding and subtracting $\CE{\Maxresidual{\SetofContexts{t+1}}{\Estimator t}}{\mathcal{G}_{t}}$
and $\frac{1}{\abs{\Psi_{t}}}\sum_{\tau\in\Psi_{t}}\Maxresidual{\SetofContexts{\tau}}{\Estimator t}$,
we only need to show, 
\[
\frac{1}{\abs{\Psi_{t}}}\sum_{\tau\in\Psi_{t}}\Maxresidual{\SetofContexts{\tau}}{\Estimator t}\le\frac{1}{\sqrt{\abs{\Psi_{t}}}}\norm{\Estimator t-\beta^{*}}_{V_{t}},
\]
for~\eqref{eq:regret_decomposition}. By the Cauchy-Schwartz inequality,
\[
\begin{split}\sum_{\tau\in\Psi_{t}}\Maxresidual{\SetofContexts{\tau}}{\Estimator t} & \le\sqrt{\abs{\Psi_{t}}}\sqrt{\sum_{\tau\in\Psi_{t}}\left\{ \Maxresidual{\SetofContexts{\tau}}{\Estimator t}\right\} ^{2}}\\
 & =\sqrt{\abs{\Psi_{t}}}\sqrt{\sum_{\tau\in\Psi_{t}}\max_{i\in[N]}\left\{ \Context i{\tau}^{T}\left(\Estimator t-\beta^{*}\right)\right\} ^{2}}\\
 & \le\sqrt{\abs{\Psi_{t}}}\sqrt{\sum_{\tau\in\Psi_{t}}\sum_{i=1}^{N}\left\{ \Context i{\tau}^{T}\left(\Estimator t-\beta^{*}\right)\right\} ^{2}}\\
 & \le\sqrt{\abs{\Psi_{t}}}\sqrt{\left(\Estimator t\!-\!\beta^{*}\right)^{T}\!V_{t}\left(\Estimator t\!-\!\beta^{*}\right)},
\end{split}
\]
where the last inequality holds with the fact that $V_{t}\succeq\sum_{\tau\in\Psi_{t}}\sum_{i=1}^{N}\Context i{\tau}\Context i{\tau}^{T}$. 
\end{proof}

\subsection{Proof of Lemma \ref{lem:expectation_bound}}
\begin{proof}
Let us fix $t\in[T]$ and $\Psi_{t}\subseteq[t]$. By Assumption~\ref{assum:iid_context},
$\SetofContexts t$ is independent with $\mathcal{G}_{t-1}$. Thus,
\[
\CE{\Maxresidual{\SetofContexts t}{\beta_{1}}}{\mathcal{G}_{t-1}}=\Expectation_{X}\left[\Maxresidual X{\beta_{1}}\right],
\]
where $X\in\Real^{d\times N}$ arises from $P_{X}$ defined in Assumption~\ref{assum:iid_context}.
For any $x>0$ and $\theta>0$, 
\begin{align*}
 & \CP{\sup_{\norm{\beta_{1}-\beta^{*}}_{2}\le L}\abs{\CE{\Maxresidual{\SetofContexts t}{\beta_{1}}}{\mathcal{G}_{t-1}}-\frac{1}{\abs{\Psi_{t}}}\sum_{\tau\in\Psi_{t}}\Maxresidual{\SetofContexts t}{\beta_{1}}}>x}{\Psi_{t}}\\
 & \le\exp\left(-\theta x\right)\CE{\exp\left(\theta\sup_{\norm{\beta_{1}-\beta^{*}}_{2}\le L}\abs{\Expectation_{X}\left[\Maxresidual{\SetofContexts t}{\beta_{1}}\right]-\frac{1}{\abs{\Psi_{t}}}\sum_{\tau\in\Psi_{t}}\Maxresidual{\SetofContexts t}{\beta_{1}}}\right)}{\Psi_{t}}.
\end{align*}
Let $\tau_{1}\le\tau_{2},\ldots\le\tau_{\abs{\Psi_{t}}}$ be an ordered
round in $\Psi_{t}$. Then by Assumption 3, $\SetofContexts{\tau_{1}},\ldots,\SetofContexts{\tau_{\abs{\Psi_{t}}}}$
are IID random variables and we can use the symmetrization lemma \citep[Lemma 2.3.1]{vanderVaart1996}
to have 
\begin{equation}
\begin{split}\Expectation & \left[\exp\left(\theta\sup_{\norm{\beta_{1}-\beta^{*}}_{2}\le L}\abs{\Expectation_{X}\left[\Maxresidual{\SetofContexts t}{\beta_{1}}\right]-\frac{1}{\abs{\Psi_{t}}}\sum_{\tau\in\Psi_{t}}\Maxresidual{\SetofContexts t}{\beta_{1}}}\right)\right]\\
 & \le\Expectation\left[\exp\left(2\theta\sup_{\norm{\beta_{1}-\beta^{*}}_{2}\le L}\abs{\frac{1}{\abs{\Psi_{t}}}\sum_{n=1}^{\abs{\Psi_{t}}}\xi_{n}\Maxresidual{\SetofContexts{\tau_{n}}}{\beta_{1}}}\right)\right],
\end{split}
\label{eq:expectation_exp_bound}
\end{equation}
where $\xi_{1},\ldots,\xi_{\abs{\Psi_{t}}}$ are independent Rademacher
random variables. For any $\epsilon>0$ let $\tilde{\beta}_{1},\ldots,\tilde{\beta}_{\Theta(\epsilon)}$
be the $\epsilon$-cover of $\mathcal{B}:=\left\{ \beta_{1}\in\Real^{d}:\norm{\beta_{1}-\beta^{*}}_{2}\le L\right\} $.
By the definition of $\epsilon$-cover, for each $\beta_{1}\in\mathcal{B}$,
there exists $\tilde{\beta}_{j}$ such that $\norm{\tilde{\beta}_{j}-\beta_{1}}_{2}\le\epsilon$.
Thus, 
\begin{align*}
\abs{\sum_{n=1}^{\abs{\Psi_{t}}}\xi_{n}\Maxresidual{\SetofContexts{\tau_{n}}}{\beta_{1}}}\le & \abs{\sum_{n=1}^{\abs{\Psi_{t}}}\xi_{n}\left\{ \Maxresidual{\SetofContexts{\tau_{n}}}{\beta_{1}}-\Maxresidual{\SetofContexts{\tau_{n}}}{\tilde{\beta}_{j}}\right\} }+\abs{\sum_{n=1}^{\abs{\Psi_{t}}}\xi_{n}\Maxresidual{\SetofContexts{\tau_{n}}}{\tilde{\beta}_{j}}}\\
\le & \sum_{n=1}^{\abs{\Psi_{t}}}\abs{\Maxresidual{\SetofContexts{\tau_{n}}}{\beta_{1}}-\Maxresidual{\SetofContexts{\tau_{n}}}{\tilde{\beta}_{j}}}+\abs{\sum_{n=1}^{\abs{\Psi_{t}}}\xi_{n}\Maxresidual{\SetofContexts{\tau_{n}}}{\tilde{\beta}_{j}}}.
\end{align*}
By the definition of $\Maxresidual{\SetofContexts{\tau_{n}}}{\beta_{1}}$
and Assumption 1, 
\begin{align*}
\abs{\Maxresidual{\SetofContexts{\tau_{n}}}{\beta_{1}}-\Maxresidual{\SetofContexts{\tau_{n}}}{\tilde{\beta}_{j}}}= & \abs{\max_{i}\abs{\Context i{\tau_{n}}^{T}\left(\beta^{*}-\beta_{1}\right)}-\max_{i}\abs{\Context i{\tau_{n}}^{T}\left(\beta^{*}-\tilde{\beta}_{j}\right)}}\\
\le & \max_{i}\abs{\abs{\Context i{\tau_{n}}^{T}\left(\beta^{*}-\beta_{1}\right)}-\abs{\Context i{\tau_{n}}^{T}\left(\beta^{*}-\tilde{\beta}_{j}\right)}}\\
\le & \max_{i}\abs{\Context i{\tau_{n}}^{T}\left(\beta_{1}-\tilde{\beta}_{j}\right)}\\
\le & \max_{i}\norm{\Context i{\tau_{n}}}_{2}\norm{\beta_{1}-\tilde{\beta}_{j}}_{2}\\
\le & \epsilon.
\end{align*}
Thus, 
\[
\sup_{\norm{\beta_{1}-\beta^{*}}_{2}\le L}\abs{\sum_{n=1}^{\abs{\Psi_{t}}}\xi_{n}\Maxresidual{\SetofContexts{\tau_{n}}}{\beta_{1}}}\le\abs{\Psi_{t}}\epsilon+\sup_{j=1,\ldots,\Theta(\epsilon)}\abs{\sum_{n=1}^{\abs{\Psi_{t}}}\xi_{n}\Maxresidual{\SetofContexts{\tau_{n}}}{\tilde{\beta}_{j}}}.
\]
Plugging in \eqref{eq:expectation_exp_bound} gives 
\begin{align*}
 & \CP{\sup_{\norm{\beta_{1}-\beta^{*}}_{2}\le L}\abs{\CE{\Maxresidual{\SetofContexts t}{\beta_{1}}}{\mathcal{G}_{t-1}}-\frac{1}{\abs{\Psi_{t}}}\sum_{\tau\in\Psi_{t}}\Maxresidual{\SetofContexts{\tau}}{\beta_{1}}}>x}{\Psi_{t}}\\
 & \le\exp\left(-\theta x+\theta\epsilon\right)\CE{\exp\left(\frac{2\theta}{\abs{\Psi_{t}}}\sup_{j=1,\ldots,\Theta(\epsilon)}\abs{\sum_{n=1}^{\abs{\Psi_{t}}}\xi_{n}\Maxresidual{\SetofContexts{\tau_{n}}}{\tilde{\beta}_{j}}}\right)}{\Psi_{t}}\\
 & \le\exp\left(-\theta x+\theta\epsilon\right)\sum_{j=1}^{\Theta(\epsilon)}\CE{\exp\left(\frac{2\theta}{\abs{\Psi_{t}}}\abs{\sum_{n=1}^{\abs{\Psi_{t}}}\xi_{n}\Maxresidual{\SetofContexts{\tau_{n}}}{\tilde{\beta}_{j}}}\right)}{\Psi_{t}}.
\end{align*}
Observe that for each $j=1,\ldots,\Theta(\epsilon)$, 
\[
\abs{\Maxresidual{\SetofContexts{\tau_{n}}}{\tilde{\beta}_{j}}}\le\max_{i}\norm{\Context i{\tau_{n}}}_{2}\norm{\beta^{*}-\tilde{\beta}_{j}}_{2}\le L.
\]
Then by Hoeffding's Lemma, 
\begin{align*}
 & \CE{\exp\left(\frac{2\theta}{\abs{\Psi_{t}}}\abs{\sum_{n=1}^{\abs{\Psi_{t}}}\xi_{n}\Maxresidual{\SetofContexts{\tau_{n}}}{\tilde{\beta}_{j}}}\right)}{\Psi_{t}}\\
 & =\Expectation\CE{\exp\left(\frac{2\theta}{\abs{\Psi_{t}}}\abs{\sum_{n=1}^{\abs{\Psi_{t}}}\xi_{n}\Maxresidual{\SetofContexts{\tau_{n}}}{\tilde{\beta}_{j}}}\right)}{\left\{ X(\tau_{n})\right\} _{n=1}^{\abs{\Psi_{t}}},\Psi_{t}}\\
 & =\Expectation\prod_{n=1}^{\abs{\Psi_{t}}}\CE{\exp\left(\frac{2\theta}{\abs{\Psi_{t}}}\xi_{n}\Maxresidual{\SetofContexts{\tau_{n}}}{\tilde{\beta}_{j}}\right)}{\left\{ X(\tau_{n})\right\} _{n=1}^{\abs{\Psi_{t}}},\Psi_{t}}\\
 & \le\exp\left(\frac{2\theta^{2}L^{2}}{\abs{\Psi_{t}}}\right).
\end{align*}
Thus, 
\begin{align*}
 & \CP{\sup_{\norm{\beta_{1}-\beta^{*}}_{2}\le L}\abs{\CE{\Maxresidual{\SetofContexts t}{\beta_{1}}}{\mathcal{G}_{t-1}}-\frac{1}{\abs{\Psi_{t}}}\sum_{\tau\in\Psi_{t}}\Maxresidual{\SetofContexts{\tau}}{\beta_{1}}}>x}{\Psi_{t}}\\
 & \le\exp\left(-\theta x+\theta\epsilon\right)2\Theta(\epsilon)\exp\left(\frac{2\theta^{2}L^{2}}{\abs{\Psi_{t}}}\right)\\
 & =2\Theta(\epsilon)\exp\left\{ -\theta\left(x-\epsilon\right)+\frac{2\theta^{2}L^{2}}{\abs{\Psi_{t}}}\right\} .
\end{align*}
Minimizing with respect to $\theta>0$ gives, 
\begin{align*}
 & \CP{\sup_{\norm{\beta_{1}-\beta^{*}}_{2}\le L}\abs{\CE{\Maxresidual{\SetofContexts t}{\beta_{1}}}{\mathcal{G}_{t-1}}-\frac{1}{\abs{\Psi_{t}}}\sum_{\tau\in\Psi_{t}}\Maxresidual{\SetofContexts{\tau}}{\beta_{1}}}>x}{\Psi_{t}}\\
 & \le2\Theta(\epsilon)\exp\left\{ -\frac{\abs{\Psi_{t}}\left(x-\epsilon\right)^{2}}{8L^{2}}\right\} .
\end{align*}
The covering number of $\mathcal{B}$ is bounded by $\Theta(\epsilon)\le(\frac{3L}{\epsilon})^{d}$.
Thus, with probability at least $1-\delta/T$, 
\begin{align*}
\sup_{\norm{\beta_{1}-\beta^{*}}_{2}\le L}\abs{\CE{\Maxresidual{\SetofContexts t}{\beta_{1}}}{\mathcal{G}_{t-1}}-\frac{1}{\abs{\Psi_{t}}}\sum_{\tau\in\Psi_{t}}\Maxresidual{\SetofContexts{\tau}}{\beta_{1}}} & \le\epsilon+L\sqrt{\frac{8}{\abs{\Psi_{t}}}}\sqrt{\log\frac{2\Theta(\epsilon)T}{\delta}}\\
 & \le\epsilon+L\sqrt{\frac{8}{\abs{\Psi_{t}}}}\sqrt{d\log\frac{3L}{\epsilon}+\log\frac{2T}{\delta}}.
\end{align*}
Setting $\epsilon=3L\delta/(2T)$ gives, 
\begin{align*}
\sup_{\norm{\beta_{1}-\beta^{*}}_{2}\le L}\abs{\CE{\Maxresidual{\SetofContexts t}{\beta_{1}}}{\mathcal{G}_{t-1}}-\frac{1}{\abs{\Psi_{t}}}\sum_{\tau\in\Psi_{t}}\Maxresidual{\SetofContexts{\tau}}{\beta_{1}}} & \le\frac{3L\delta}{2T}+L\sqrt{\frac{8}{\abs{\Psi_{t}}}}\sqrt{d\log\frac{2T}{\delta}+\log\frac{2T}{\delta}}\\
 & \le\frac{3L\delta}{2T}+4L\sqrt{\frac{1}{\abs{\Psi_{t}}}}\sqrt{d\log\frac{2T}{\delta}}.
\end{align*}
\end{proof}

\subsection{Proof of Theorem \ref{thm:Estimation_error}}

\subsubsection{A bound for the imputation estimator}

\label{subsec:imputation_estimator} To prove Theorem \ref{thm:Estimation_error},
we need to prove the following bound for the imputation estimator
$\Impute t$ which is used in $\Tildereward it$ and $\Estimator t$.
The proposed imputation estimator is used to obtain the bound~\eqref{eq:imputation_estimator_bound}
exploiting Assumptions~\ref{assum:boundedness}-\ref{assum:iid_context}.
\begin{lem}
Suppose Assumptions~\ref{assum:boundedness}-\ref{assum:iid_context}
hold. Let 
\begin{equation}
\begin{split}\Impute t:= & \left(\!\sum_{\tau\in\Psi_{t}}\sum_{i=1}^{N}\Context i{\tau}\Context i{\tau}^{T}\!+\!\!\sum_{\tau\notin\Psi_{t}}\Context{\Action{\tau}}{\tau}\Context{\Action{\tau}}{\tau}^{T}\!+\!\gamma_{t}I\!\right)^{-1}\\
 & \left\{ \sum_{\tau\in\Psi_{t}}\sum_{i=1}^{N}\Context i{\tau}\left(\left\{ 1-\frac{\Indicator{\Tildeaction{\tau}=i}}{\SelectionP i{\tau}}\right\} \Context i{\tau}^{T}\Ridgebeta{t-1}+\frac{\Indicator{\Tildeaction{\tau}=i}}{\SelectionP i{\tau}}Y_{\Tildeaction t,t}\right)+\sum_{\tau\notin\Psi_{t}}\Context{\Action{\tau}}{\tau}\Reward{\tau}\right\} ,
\end{split}
\label{eq:imputation_estimator}
\end{equation}
for $\gamma_{t}:=4\sqrt{2}N\sqrt{\abs{\Psi_{t}}\log\frac{4t^{2}}{\delta}}$
and $\Ridgebeta t$ is a normalized ridge estimator using pairs of
selected contexts and corresponding rewards until round t, i.e. 
\[
\Ridgebeta t:=\frac{\left(\sum_{\tau=1}^{t}\Context{\Action{\tau}}{\tau}\Context{\Action{\tau}}{\tau}^{T}+I_{d}\right)^{-1}\left(\sum_{\tau=1}^{t}\Context{\Action{\tau}}{\tau}\Reward{\tau}\right)}{\max\left\{ \norm{\left(\sum_{\tau=1}^{t}\Context{\Action{\tau}}{\tau}\Context{\Action{\tau}}{\tau}^{T}+I_{d}\right)^{-1}\left(\sum_{\tau=1}^{t}\Context{\Action{\tau}}{\tau}\Reward{\tau}\right)},1\right\} }.
\]
Then with probability at least $1-\delta$, 
\begin{equation}
\norm{\Impute t-\beta^{*}}_{2}\le\frac{1}{N},\label{eq:imputation_estimator_bound}
\end{equation}
holds for $t\ge\max\left\{ \frac{8}{p}\log\frac{4T}{\delta},C_{p,\sigma}N^{2}\phi^{-4}\log\frac{8T}{\delta}\right\} $.
\label{lem:imputation_estimator_bound} 
\end{lem}

\begin{rem}
In deriving the bound~\eqref{eq:imputation_estimator_bound}, the
minimum eigenvalue of the Gram matrix is required to be $\Omega(t)$,
which is challenging even under Assumption~\ref{assum:iid_context}
when the ridge estimator consist of only selected contexts and rewards
is used (See Section 5 in \citep{li2017provably}). Therefore we propose
the imputation estimator as in~\ref{eq:imputation_estimator} which
uses the contexts from all arms to exploit Assumption~\ref{assum:iid_context}
elevating the minimum eigenvalue of the Gram matrix. 
\end{rem}

\begin{proof}
{[}Step 1. Bounding the minimum eigenvalue of the Gram matrix{]} Fix
$t$ and set 
\begin{align*}
\gamma_{t} & :=4\sqrt{2}N\sqrt{\abs{\Psi_{t}}\log\frac{4t^{2}}{\delta}},\\
W_{t} & :=\sum_{\tau\in\Psi_{t}}\sum_{i=1}^{N}\Context i{\tau}\Context i{\tau}^{T}+\sum_{\tau\notin\Psi_{t}}\Context{\Action{\tau}}{\tau}\Context{\Action{\tau}}{\tau}+\gamma_{t}I.
\end{align*}
Then by definition of $\Impute t$, we have 
\begin{equation}
\begin{split}\norm{\Impute t-\beta^{*}}_{2}= & \norm{W_{t}^{-1}\left(\sum_{\tau\in\Psi_{t}}\sum_{i=1}^{N}\Context i{\tau}\Tildereward i{\tau}+\sum_{\tau\notin\Psi_{t}}\Context{\Action{\tau}}{\tau}\Reward{\tau}-W_{t}\beta^{*}\right)}_{2}\\
\le & \norm{W_{t}^{-1}}_{2}\left\{ \norm{\sum_{\tau\in\Psi_{t}}\sum_{i=1}^{N}\Context i{\tau}\left(\Tildereward i{\tau}-\Context i{\tau}^{T}\beta^{*}\right)+\sum_{\tau\notin\Psi_{t}}\Context{\Action{\tau}}{\tau}\Error{\Action{\tau}}{\tau}}_{2}+\gamma_{t}\norm{\beta^{*}}_{2}\right\} \\
\le & \Mineigen{W_{t}}^{-1}\left\{ \norm{\sum_{\tau\in\Psi_{t}}\sum_{i=1}^{N}\Context i{\tau}\left(\Tildereward i{\tau}-\Context i{\tau}^{T}\beta^{*}\right)+\sum_{\tau\notin\Psi_{t}}\Context{\Action{\tau}}{\tau}\Error{\Action{\tau}}{\tau}}_{2}+\gamma_{t}\right\} ,
\end{split}
\label{eq:Impute_decompostion}
\end{equation}
where $\eta_{i,t}=Y_{i,t}-\Context it^{T}\beta^{*}$. For the minimum
eigenvalue term, we have 
\[
\Mineigen{W_{t}}\ge\Mineigen{\sum_{\tau\in\Psi_{t}}\sum_{i=1}^{N}\Context i{\tau}\Context i{\tau}^{T}+\gamma_{t}I_{d}}.
\]
Let $\tau_{1}<\tau_{2}<\cdots<\tau_{\abs{\Psi_{t}}}$ be the ordered
rounds in $\Psi_{t}$. Since $\norm{\sum_{i=1}^{N}\Context i{\tau}\Context i{\tau}^{T}}_{F}\le N$
and 
\[
\Mineigen{\CE{\sum_{i=1}^{N}\Context i{\tau_{k}}\Context i{\tau_{k}}^{T}}{\SetofContexts{\tau_{1}},\ldots,\SetofContexts{\tau_{k-1}}}}=\Mineigen{\Expectation\left[\sum_{i=1}^{N}\Context i{\tau_{k}}\Context i{\tau_{k}}^{T}\right]}\ge N\phi^{2},
\]
we can use Lemma 6 in \citet{kim2021doubly} to have 
\begin{equation}
\Mineigen{W_{t}}\ge\Mineigen{\sum_{\tau\in\Psi_{t}}\sum_{i=1}^{N}\Context i{\tau}\Context i{\tau}^{T}+\gamma_{t}I_{d}}\ge\abs{\Psi_{t}}N\phi^{2}.\label{eq:Impute_mineigen}
\end{equation}

{[}Step 2. Estimation error decomposition{]} By definition of $\Tildereward i{\tau}$,
we have 
\begin{align*}
\sum_{\tau\in\Psi_{t}}\sum_{i=1}^{N}\Context i{\tau}\left(\Tildereward i{\tau}-\Context i{\tau}^{T}\beta^{*}\right)= & \sum_{\tau\in\Psi_{t}}\sum_{i=1}^{N}\left(1-\frac{\Indicator{\Tildeaction{\tau}=i}}{\SelectionP i{\tau}}\right)\Context i{\tau}\Context i{\tau}^{T}\left(\Ridgebeta{t-1}-\beta^{*}\right)\\
 & +\sum_{\tau\in\Psi_{t}}\sum_{i=1}^{N}\frac{\Indicator{\Tildeaction{\tau}=i}}{\SelectionP i{\tau}}\Error i{\tau}\\
= & \sum_{\tau\in\Psi_{t}}\sum_{i=1}^{N}\left(1-\frac{\Indicator{\Tildeaction{\tau}=i}}{\SelectionP i{\tau}}\right)\XX i{\tau}\left(\Ridgebeta{t-1}-\beta^{*}\right)\\
 & +\sum_{\tau\in\Psi_{t}}\frac{\Error{\Tildeaction{\tau}}{\tau}}{\SelectionP{\Tildeaction{\tau}}{\tau}}\Context{\Tildeaction{\tau}}{\tau},
\end{align*}
where $\XX i{\tau}=\Context i{\tau}\Context i{\tau}^{T}$. Plugging
this and \eqref{eq:Impute_mineigen} in \eqref{eq:Impute_decompostion}
gives, 
\begin{equation}
\begin{split}\norm{\Impute t-\beta^{*}}_{2}%
 & \le\frac{1}{\abs{\Psi_{t}}N\phi^{2}}\norm{\sum_{\tau\in\Psi_{t}}\sum_{i=1}^{N}\left(1-\frac{\Indicator{\Tildeaction{\tau}=i}}{\SelectionP i{\tau}}\right)\XX i{\tau}\left(\Ridgebeta{t-1}-\beta^{*}\right)}_{2}\\
 & \quad+\frac{1}{\abs{\Psi_{t}}N\phi^{2}}\norm{\sum_{\tau\in\Psi_{t}}\frac{\Error{\Tildeaction{\tau}}{\tau}}{\SelectionP{\Tildeaction{\tau}}{\tau}}\Context{\Tildeaction{\tau}}{\tau}+\sum_{\tau\notin\Psi_{t}}\Error{\Action{\tau}}{\tau}\Context{\Action{\tau}}{\tau}}_{2}+\frac{4\sqrt{2\log\frac{4t^{2}}{\delta}}}{\phi^{2}\sqrt{\abs{\Psi_{t}}}}.
\end{split}
\label{eq:Impute_eta_decompostion}
\end{equation}

{[}Step 3. Bounding the first term in \eqref{eq:Impute_eta_decompostion}{]}
For the first term, 
\begin{align*}
 & \norm{\sum_{\tau\in\Psi_{t}}\sum_{i=1}^{N}\left(1-\frac{\Indicator{\Tildeaction{\tau}=i}}{\SelectionP i{\tau}}\right)\XX i{\tau}\left(\Ridgebeta{t-1}-\beta^{*}\right)}_{2}\\
\le & \norm{\sum_{\tau\in\Psi_{t}}\sum_{i=1}^{N}\left(1-\frac{\Indicator{\Tildeaction{\tau}=i}}{\SelectionP i{\tau}}\right)\XX i{\tau}}_{2}\norm{\Ridgebeta{t-1}-\beta^{*}}_{2}\\
\le & 2\norm{\sum_{\tau\in\Psi_{t}}\sum_{i=1}^{N}\left(1-\frac{\Indicator{\Tildeaction{\tau}=i}}{\SelectionP i{\tau}}\right)\XX i{\tau}}_{F}
\end{align*}
Define the filtration as $\mathcal{G}_{0}=\Psi_{t}$ and $\mathcal{G}_{\tau}=\mathcal{G}_{\tau-1}\cup\{\SetofContexts{\tau},\Tildeaction{\tau},\Action{\tau}\}$
for $\tau\in[t]$. This filtration refers to the case where the subset
of rounds $\Psi_{t}$ for using contexts from all arms is observed
first and $\Tildeaction 1,\Action 1,\Tildeaction 2,\Action 2\ldots,\Tildeaction t$
are observed later. In \texttt{HyRan Bandit}, the hybridization variables
$\Tildeaction 1,\ldots,\Tildeaction t$ and actions $\Action 1,\ldots,\Action t$
are observed first to determine $\Psi_{t}$. But in theoretical analysis,
we change the order of observation by defining a new filtration $\mathcal{G}_{0},\ldots,\mathcal{G}_{t}$
and obtain a suitable bound with the martingale method \citep{kontorovich2008concentration}.
Set 
\[
M:=\sum_{\tau\in\Psi_{t}}\sum_{i=1}^{N}\left(1-\frac{\Indicator{\Tildeaction{\tau}=i}}{\SelectionP i{\tau}}\right)\XX i{\tau},
\]
and define $M_{\tau}=\CE M{\mathcal{G}_{\tau}}$. Then $\{M_{\tau}\}_{\tau=0}^{t}$
is a $\Real^{d\times d}$-valued martingale sequence since 
\[
\CE{M_{\tau}}{\mathcal{G}_{\tau-1}}=\CE{\CE M{\mathcal{G}_{\tau}}}{\mathcal{G}_{\tau-1}}=\CE M{\mathcal{G}_{\tau-1}}=M_{\tau-1}.
\]
By Lemma \ref{lem:hilbert_to_R2}, we can find a $\Real^{2}$-valued
martingale sequence $\{N_{\tau}\}_{\tau=0}^{t}$ such that $N_{0}=(0,0)^{T}$
and 
\[
\norm{M_{\tau}}_{F}=\norm{N_{\tau}}_{2},\quad\norm{M_{\tau}-M_{\tau-1}}_{F}=\norm{N_{\tau}-N_{\tau-1}}_{2},
\]
for all $\tau\in[t]$. Set $N_{\tau}=(N_{\tau}^{(1)},N_{\tau}^{(2)})^{T}$.
Then for each $r=1,2$ and $\tau\in[t]$, 
\[
\begin{split}\abs{N_{\tau}^{(r)}-N_{\tau-1}^{(r)}}\le & \norm{N_{\tau}-N_{\tau-1}}_{2}\\
= & \norm{M_{\tau}-M_{\tau-1}}_{F}\\
= & \norm{\CE M{\mathcal{G}_{\tau}}-\CE M{\mathcal{G}_{\tau-1}}}_{F}\\
= & \begin{cases}
\norm{\sum_{i=1}^{N}\left(1-\frac{\Indicator{\Tildeaction{\tau}=i}}{\SelectionP i{\tau}}\right)\XX i{\tau}}_{F} & \tau\in\Psi_{t}\\
0 & \tau\notin\Psi_{t}
\end{cases}\\
\le & \begin{cases}
\norm{\sum_{i=1}^{N}\XX i{\tau}}_{F}+\norm{\frac{1}{\SelectionP{\Tildeaction{\tau}}{\tau}}\XX i{\tau}}_{F} & \tau\in\Psi_{t}\\
0 & \tau\notin\Psi_{t}
\end{cases}\\
\le & \begin{cases}
N\left(\frac{2-p}{1-p}\right) & \tau\in\Psi_{t}\\
0 & \tau\notin\Psi_{t}
\end{cases},
\end{split}
\]
holds almost surely. The third equality holds since for any $\tau\in[t]$,
\[
\begin{split}\CE{\sum_{i=1}^{N}\left(1-\frac{\Indicator{\Tildeaction u=i}}{\SelectionP iu}\right)}{\mathcal{G}_{\tau}} & =0,\,\forall u>\tau,\\
\CE M{\mathcal{G}_{\tau}} & =\sum_{u\in\Psi_{t},u\le\tau}\sum_{i=1}^{N}\left(1-\frac{\Indicator{\Tildeaction{\tau}=i}}{\SelectionP i{\tau}}\right)\XX i{\tau}.
\end{split}
\]
Using Lemma \ref{lem:azuma}, for $x>0$ and $r=1,2$, 
\[
\CP{\abs{N_{\tau}^{(r)}}>x}{\mathcal{G}_{0}}\le2\exp\left(-\frac{x^{2}}{2N^{2}\abs{\Psi_{t}}\left(\frac{2-p}{1-p}\right)^{2}}\right),
\]
which implies that 
\[
\CP{\abs{N_{\tau}^{(r)}}>N\left(\frac{2-p}{1-p}\right)\sqrt{2\abs{\Psi_{t}}\log\frac{4t^{2}}{\delta}}}{\mathcal{G}_{0}}\le\frac{\delta}{2t^{2}}.
\]
Since 
\[
\norm M_{F}=\norm{M_{t}}_{F}=\norm{N_{t}}_{2}\le\abs{N_{t}^{(1)}}+\abs{N_{t}^{(2)}},
\]
we have 
\[
\CP{\norm M_{F}>2N\left(\frac{2-p}{1-p}\right)\sqrt{2\abs{\Psi_{t}}\log\frac{4t^{2}}{\delta}}}{\mathcal{G}_{0}}\le\frac{\delta}{t^{2}},
\]
for any subset $\Psi_{t}\subseteq[t]$. Thus, we conclude that 
\[
\begin{split}\Probability & \left(\norm{\sum_{\tau\in\Psi_{t}}\sum_{i=1}^{N}\left(1-\frac{\Indicator{\Tildeaction{\tau}=i}}{\SelectionP i{\tau}}\right)\XX i{\tau}\left(\Impute{t-1}-\beta^{*}\right)}_{2}>4N\left(\frac{2-p}{1-p}\right)\sqrt{2\abs{\Psi_{t}}\log\frac{4t^{2}}{\delta}}\right)\\
\le & \Probability\left(2\norm M_{F}>4N\left(\frac{2-p}{1-p}\right)\sqrt{2\abs{\Psi_{t}}\log\frac{4t^{2}}{\delta}}\right)\\
\le & \Expectation\CP{2\norm M_{F}>4N\left(\frac{2-p}{1-p}\right)\sqrt{2\abs{\Psi_{t}}\log\frac{4t^{2}}{\delta}}}{\Psi_{t}}\\
\ \le & \frac{\delta}{t^{2}}.
\end{split}
\]

{[}Step 4. Bounding the second term in \eqref{eq:Impute_eta_decompostion}{]}
Now for the second term in \eqref{eq:Impute_eta_decompostion}, we
have for any $x>0$, 
\[
\begin{split}\Probability & \left(\norm{\sum_{\tau\in\Psi_{t}}\frac{\Error{\Tildeaction{\tau}}{\tau}}{\SelectionP{\Tildeaction{\tau}}{\tau}}\Context{\Tildeaction{\tau}}{\tau}+\sum_{\tau\notin\Psi_{t}}\Error{\Action{\tau}}{\tau}\Context{\Action{\tau}}{\tau}}_{2}>x\right)\\
\le & \Probability\left(\left\{ \norm{\sum_{\tau\in\Psi_{t}}\frac{\Error{\Tildeaction{\tau}}{\tau}}{\SelectionP{\Tildeaction{\tau}}{\tau}}\Context{\Tildeaction{\tau}}{\tau}+\sum_{\tau\notin\Psi_{t}}\Error{\Action{\tau}}{\tau}\Context{\Action{\tau}}{\tau}}_{2}>x\right\} \bigcap\left\{ \bigcap_{\tau\in\Psi_{t}}\left\{ \Tildeaction{\tau}=\Action{\tau}\right\} \right\} \right)\\
 & +\Probability\left(\bigcup_{\tau\in\Psi_{t}}\left\{ \Tildeaction{\tau}\neq\Action{\tau}\right\} \right)\\
\le & \Probability\left(\norm{\sum_{\tau\in\Psi_{t}}\frac{\Error{\Action{\tau}}{\tau}}{\SelectionP{\Action{\tau}}{\tau}}\Context{\Action{\tau}}{\tau}+\sum_{\tau\notin\Psi_{t}}\Error{\Action{\tau}}{\tau}\Context{\Action{\tau}}{\tau}}_{2}>x\right).
\end{split}
\]
The last inequality holds since \texttt{HyRan Bandit} selects allocates
the round $\tau$ in $\Psi_{t}$ only when $\Tildeaction{\tau}=\Action{\tau}$,
almost surely. Since $\SelectionP{\Action{\tau}}{\tau}=p$, we observe
that $\frac{\Error{\Action{\tau}}{\tau}}{\SelectionP{\Action{\tau}}{\tau}}$
and $\Error{\Action{\tau}}{\tau}$ are $\frac{\sigma}{p}$-sub-Gaussian.
Using Lemma 4 in \citet{kim2021doubly} we have, 
\[
\Probability\left(\norm{\sum_{\tau\in\Psi_{t}}\frac{\Error{\Action{\tau}}{\tau}}{\SelectionP{\Action{\tau}}{\tau}}\Context{\Action{\tau}}{\tau}+\sum_{\tau\notin\Psi_{t}}\Error{\Action{\tau}}{\tau}\Context{\Action{\tau}}{\tau}}_{2}>\frac{C\sigma}{p}\sqrt{t}\sqrt{\log\frac{4t^{2}}{\delta}}\right)\le\frac{\delta}{t^{2}},
\]
for some absolute constant $C>0$. Now from \eqref{eq:Impute_eta_decompostion},
with probability $1-\frac{3\delta}{t^{2}}$, we have 
\[
\norm{\Impute t-\beta^{*}}_{2}\le\frac{1}{\abs{\Psi_{t}}N\phi^{2}}\left\{ 4N\left(\frac{2-p}{1-p}\right)\sqrt{2\abs{\Psi_{t}}\log\frac{4t^{2}}{\delta}}+\frac{C\sigma}{p}\sqrt{t}\sqrt{\log\frac{4t^{2}}{\delta}}\right\} +\frac{4\sqrt{2\log\frac{4t^{2}}{\delta}}}{\phi^{2}\sqrt{\abs{\Psi_{t}}}}.
\]
By Lemma \ref{lem:psi_size}, $\abs{\Psi_{t}}\ge\frac{p}{2}t$ for
all $t\ge\frac{1}{p}\log\frac{T}{\delta}$, with probability at least
$1-\delta$. Then we have 
\[
\begin{split}\norm{\Impute t-\beta^{*}}_{2}\le & \frac{1}{\phi^{2}\sqrt{t}}\left\{ \frac{8(2-p)}{(1-p)\sqrt{p}}+\frac{\sqrt{2}C\sigma}{p^{2}N}+\frac{8}{\sqrt{p}}\right\} \sqrt{2\log\frac{4t^{2}}{\delta}}\\
\le & \frac{2}{\phi^{2}\sqrt{t}}\left\{ \frac{8(2-p)}{(1-p)\sqrt{p}}+\frac{\sqrt{2}C\sigma}{p^{2}}+\frac{8}{\sqrt{p}}\right\} \sqrt{\log\frac{2T}{\delta}}.
\end{split}
\]
Set 
\begin{equation}
C_{p,\sigma}:=\frac{8(2-p)}{(1-p)\sqrt{p}}+\frac{\sqrt{2}C\sigma}{p^{2}}+\frac{8}{\sqrt{p}}.\label{eq:C_p_sigma}
\end{equation}
Then for all $t\ge\max\left\{ \frac{1}{p}\log\frac{T}{\delta},C_{p,\sigma}N^{2}\phi^{-4}\log\frac{2T}{\delta}\right\} $,
we have $\norm{\Impute t-\beta^{*}}_{2}\le\frac{1}{N}$, with probability
at least $1-4\delta$. 
\end{proof}

\subsubsection{Proof of Theorem \ref{thm:Estimation_error}}

Now we are ready to prove Theorem \ref{thm:Estimation_error}. 
\begin{proof}
{[}Step 1. Decompostion{]} By the definition of $\Estimator t$ in
\ref{eq:estimator}, 
\begin{align*}
\norm{\Estimator t-\beta^{*}}_{V_{t}}= & \norm{V_{t}^{-1}\left(\sum_{\tau\in\Psi_{t}}\sum_{i=1}^{N}\Context i{\tau}\Tildereward i{\tau}+\sum_{\tau\notin\Psi_{t}}\Context{\Action{\tau}}{\tau}Y_{\Action{\tau},\tau}V_{t}\beta^{*}\right)}_{V_{t}}\\
= & \norm{\sum_{\tau\in\Psi_{t}}\sum_{i=1}^{N}\Context i{\tau}\Tildereward i{\tau}+\sum_{\tau\notin\Psi_{t}}\Context{\Action{\tau}}{\tau}Y_{\Action{\tau},\tau}V_{t}\beta^{*}}_{V_{t}^{-1}}\\
= & \norm{\sum_{\tau\in\Psi_{t}}\sum_{i=1}^{N}\Context i{\tau}\left(\Tildereward i{\tau}-\Context i{\tau}^{T}\beta^{*}\right)+\sum_{\tau\notin\Psi_{t}}\Context{\Action{\tau}}{\tau}\left(Y_{\Action{\tau},\tau}-\Context{\Action{\tau}}{\tau}^{T}\beta^{*}\right)-\lambda_{t}\beta^{*}}_{V_{t}^{-1}}.
\end{align*}
Set $\TildeError i{\tau}:=\Tildereward i{\tau}-\Context i{\tau}^{T}\beta^{*}$.
Since $Y_{\Action{\tau},\tau}=\Context{\Action{\tau}}{\tau}^{T}\beta^{*}+\Error{\Action{\tau}}{\tau}$,
we have 
\begin{equation}
\begin{split}\norm{\Estimator t-\beta^{*}}_{V_{t}}= & \norm{\sum_{\tau\in\Psi_{t}}\sum_{i=1}^{N}\TildeError i{\tau}\Context i{\tau}+\sum_{\tau\notin\Psi_{t}}\Error{\Action{\tau}}{\tau}\Context{\Action{\tau}}{\tau}-\lambda_{t}\beta^{*}}_{V_{t}^{-1}}\\
\le & \norm{\lambda_{t}\beta^{*}}_{V_{t}^{-1}}+\norm{\sum_{\tau\in\Psi_{t}}\sum_{i=1}^{N}\TildeError i{\tau}\Context i{\tau}+\sum_{\tau\notin\Psi_{t}}\Error{\Action{\tau}}{\tau}\Context{\Action{\tau}}{\tau}}_{V_{t}^{-1}}.
\end{split}
\label{eq:estimator_decompose}
\end{equation}
For the first term, we have 
\begin{equation}
\norm{\lambda_{t}\beta^{*}}_{V_{t}^{-1}}\le\sqrt{\Maxeigen{V_{t}^{-1}}}\norm{\lambda_{t}\beta^{*}}_{2}\le\sqrt{\lambda_{t}}\norm{\beta^{*}}_{2}\le\sqrt{\lambda_{t}},\label{eq:estimator_last_term}
\end{equation}
where the last inequality holds due to Assumption 1. For the second
term, we use the decomposition, 
\begin{align*}
\sum_{\tau\in\Psi_{t}}\sum_{i=1}^{N}\TildeError i{\tau}\Context i{\tau}= & \sum_{\tau\in\Psi_{t}}\sum_{i=1}^{N}\left(1-\frac{\Indicator{\Tildeaction{\tau}=i}}{\SelectionP i{\tau}}\right)\Context i{\tau}\Context i{\tau}^{T}(\Impute t-\beta^{*})\\
 & +\sum_{\tau\in\Psi_{t}}\sum_{i=1}^{N}\frac{\Indicator{\Tildeaction{\tau}=i}}{\SelectionP i{\tau}}\Error i{\tau}\Context i{\tau},
\end{align*}
to have 
\begin{equation}
\begin{split} & \norm{\sum_{\tau\in\Psi_{t}}\sum_{i=1}^{N}\TildeError i{\tau}\Context i{\tau}+\sum_{\tau\notin\Psi_{t}}\Error{\Action{\tau}}{\tau}\Context{\Action{\tau}}{\tau}}_{V_{t}^{-1}}\\
 & \le\norm{\sum_{\tau\in\Psi_{t}}\sum_{i=1}^{N}\left(1-\frac{\Indicator{\Tildeaction{\tau}=i}}{\SelectionP i{\tau}}\right)\Context i{\tau}\Context i{\tau}^{T}(\Impute t-\beta^{*})}_{V_{t}^{-1}}\\
 & \quad+\norm{\sum_{\tau\in\Psi_{t}}\sum_{i=1}^{N}\frac{\Indicator{\Tildeaction{\tau}=i}}{\SelectionP i{\tau}}\Error i{\tau}\Context i{\tau}+\sum_{\tau\notin\Psi_{t}}\Error{\Action{\tau}}{\tau}\Context{\Action{\tau}}{\tau}}_{V_{t}^{-1}}.
\end{split}
\label{eq:estimator_norm_decomposition}
\end{equation}

{[}Step 2. Bounding the first term in \eqref{eq:estimator_norm_decomposition}{]}
Let $\XX i{\tau}:=\Context i{\tau}\Context i{\tau}^{T}$. For the
first term, we can use Lemma \ref{lem:imputation_estimator_bound}
to have 
\begin{align*}
 & \norm{\sum_{\tau\in\Psi_{t}}\sum_{i=1}^{N}\left(1-\frac{\Indicator{\Tildeaction{\tau}=i}}{\SelectionP i{\tau}}\right)\XX i{\tau}(\Impute t-\beta^{*})}_{V_{t}^{-1}}\\
 & =\norm{\sum_{\tau\in\Psi_{t}}\sum_{i=1}^{N}\left(1-\frac{\Indicator{\Tildeaction{\tau}=i}}{\SelectionP i{\tau}}\right)V_{t}^{-\frac{1}{2}}\XX i{\tau}(\Impute t-\beta^{*})}_{2}\\
 & \le\norm{\sum_{\tau\in\Psi_{t}}\sum_{i=1}^{N}\left(1-\frac{\Indicator{\Tildeaction{\tau}=i}}{\SelectionP i{\tau}}\right)V_{t}^{-\frac{1}{2}}\XX i{\tau}}_{2}\norm{\Impute t-\beta^{*}}_{2}\\
 & \le\frac{1}{N}\norm{\sum_{\tau\in\Psi_{t}}\sum_{i=1}^{N}\left(1-\frac{\Indicator{\Tildeaction{\tau}=i}}{\SelectionP i{\tau}}\right)V_{t}^{-\frac{1}{2}}\XX i{\tau}}_{F}.
\end{align*}
With similar technique in the proof of Lemma \ref{lem:imputation_estimator_bound},
define the filtration as $\mathcal{G}_{0}=\Psi_{t}\cup\{\SetofContexts 1,\ldots,\SetofContexts t\}$
and $\mathcal{G}_{\tau}=\mathcal{G}_{\tau-1}\cup\{\Tildeaction{\tau},\Action{\tau}\}$
for $\tau\in[t]$. Set 
\[
M:=\sum_{\tau\in\Psi_{t}}\sum_{i=1}^{N}\left(1-\frac{\Indicator{\Tildeaction{\tau}=i}}{\SelectionP i{\tau}}\right)V_{t}^{-\frac{1}{2}}\XX i{\tau},
\]
and define $M_{\tau}=\CE M{\mathcal{G}_{\tau}}$. Then $\{M_{\tau}\}_{\tau=0}^{t}$
is a $\Real^{d\times d}$-valued martingale sequence. Since for any
$\tau\in[t]$, the contexts $\SetofContexts{\tau+1},\ldots,\SetofContexts t$
are independent of $\Tildeaction{\tau}$ and 
\[
\CE{\sum_{i=1}^{N}\left(1-\frac{\Indicator{\Tildeaction u=i}}{\SelectionP iu}\right)V_{t}^{-\frac{1}{2}}\XX iu}{\mathcal{G}_{\tau}}=V_{t}^{-\frac{1}{2}}\sum_{i=1}^{N}\CE{1-\frac{\Indicator{\Tildeaction u=i}}{\SelectionP iu}}{\mathcal{G}_{\tau}}\XX iu=0,
\]
for all $u>\tau$. This leads to 
\[
\CE M{\mathcal{G}_{\tau}}=\sum_{u\in\Psi_{t},u\le\tau}\sum_{i=1}^{N}\left(1-\frac{\Indicator{\Tildeaction{\tau}=i}}{\SelectionP i{\tau}}\right)V_{t}^{-\frac{1}{2}}\XX i{\tau}.
\]
By Lemma \ref{lem:hilbert_to_R2}, we can find a $\Real^{2}$-valued
martingale sequence $\{N_{\tau}\}_{\tau=0}^{t}$ such that $N_{0}=(0,0)^{T}$
and 
\[
\norm{M_{\tau}}_{F}=\norm{N_{\tau}}_{2},\quad\norm{M_{\tau}-M_{\tau-1}}_{F}=\norm{N_{\tau}-N_{\tau-1}}_{2},
\]
for all $\tau\in[t]$. Set $N_{\tau}=(N_{\tau}^{(1)},N_{\tau}^{(2)})^{T}$.
Then for each $r=1,2$ and $\tau\in[t]$, 
\begin{align*}
\abs{N_{\tau}^{(r)}-N_{\tau-1}^{(r)}}\le & \norm{N_{\tau}-N_{\tau-1}}_{2}\\
= & \norm{M_{\tau}-M_{\tau-1}}_{F}\\
= & \norm{\CE M{\mathcal{G}_{\tau}}-\CE M{\mathcal{G}_{\tau-1}}}_{F}\\
= & \begin{cases}
\norm{\sum_{i=1}^{N}\left(1-\frac{\Indicator{\Tildeaction{\tau}=i}}{\SelectionP i{\tau}}\right)V_{t}^{-\frac{1}{2}}\XX i{\tau}}_{F} & \tau\in\Psi_{t}\\
0 & \tau\notin\Psi_{t}
\end{cases}\\
\le & \begin{cases}
\sqrt{\sum_{i=1}^{N}\left(1-\frac{\Indicator{\Tildeaction{\tau}=i}}{\SelectionP i{\tau}}\right)^{2}}\sqrt{\sum_{i=1}^{N}\norm{V_{t}^{-\frac{1}{2}}\XX i{\tau}}_{F}^{2}} & \tau\in\Psi_{t}\\
0 & \tau\notin\Psi_{t}
\end{cases}\\
\le & \begin{cases}
2\frac{N}{1-p}\sqrt{\sum_{i=1}^{N}\norm{\Context i{\tau}}_{V_{t}^{-1}}^{2}} & \tau\in\Psi_{t}\\
0 & \tau\notin\Psi_{t}
\end{cases},
\end{align*}
holds almost surely. The last inequality holds due to 
\begin{align*}
\norm{V_{t}^{-1/2}\XX i{\tau}}_{F}^{2}= & \Trace{\XX i{\tau}^{T}V_{t}^{-1}\XX i{\tau}}\\
= & \Context i{\tau}^{T}V_{t}^{-1}\Context i{\tau}\Trace{\Context i{\tau}\Context i{\tau}^{T}}\\
= & \norm{\Context i{\tau}}_{V_{t}^{-1}}^{2}\norm{\Context i{\tau}}_{2}\\
\le & \norm{\Context i{\tau}}_{V_{t}^{-1}}^{2}.
\end{align*}
Using Lemma \ref{lem:azuma}, for $x>0$ and $r=1,2$, 
\[
\CP{\abs{N_{\tau}^{(r)}}>x}{\mathcal{G}_{0}}\le2\exp\left\{ -\frac{x^{2}}{2\left(\frac{2N}{1-p}\right)^{2}\sum_{\tau\in\Psi_{t}}\sum_{i=1}^{N}\norm{\Context i{\tau}}_{V_{t}^{-1}}^{2}}\right\} ,
\]
which implies that 
\[
\CP{\abs{N_{\tau}^{(r)}}>\frac{2N}{1-p}\sqrt{2\left(\sum_{\tau\in\Psi_{t}}\sum_{i=1}^{N}\norm{\Context i{\tau}}_{V_{t}^{-1}}^{2}\right)\log\frac{4t^{2}}{\delta}}}{\mathcal{G}_{0}}\le\frac{\delta}{2t^{2}}.
\]
Since 
\[
\norm M_{F}=\norm{M_{t}}_{F}=\norm{N_{t}}_{2}\le\abs{N_{t}^{(1)}}+\abs{N_{t}^{(2)}},
\]
we have 
\[
\CP{\norm M_{F}>\frac{4N}{1-p}\sqrt{2\left(\sum_{\tau\in\Psi_{t}}\sum_{i=1}^{N}\norm{\Context i{\tau}}_{V_{t}^{-1}}^{2}\right)\log\frac{4t^{2}}{\delta}}}{\Psi_{t}}\le\frac{\delta}{t^{2}},
\]
for any subset $\Psi_{t}\subseteq[t]$. Let $U_{t}:=\sum_{\tau\in\Psi_{t}}\sum_{i=1}^{N}\Context i{\tau}\Context i{\tau}^{T}+\lambda_{t}I$.
Since $V_{t}\succeq U_{t}$, we have $\norm{\Context i{\tau^{(u)}}}_{V_{t}^{-1}}^{2}\le\norm{\Context i{\tau^{(u)}}}_{U_{t}^{-1}}^{2}$.
By the definition of the Frobenous norm and $\XX i{\tau}$, we have
\begin{align*}
\sum_{\tau\in\Psi_{t}}\sum_{i=1}^{N}\norm{\Context i{\tau^{(u)}}}_{U_{t}^{-1}}^{2}= & \sum_{\tau\in\Psi_{t}}\sum_{i=1}^{N}\Context i{\tau}^{T}U_{t}^{-1}\Context i{\tau}\\
= & \sum_{\tau\in\Psi_{t}}\sum_{i=1}^{N}\Trace{\Context i{\tau}^{T}U_{t}^{-1}\Context i{\tau}}\\
= & \sum_{\tau\in\Psi_{t}}\sum_{i=1}^{N}\Trace{\Context i{\tau}\Context i{\tau}^{T}U_{t}^{-1}}\\
= & \Trace{\left(\sum_{\tau\in\Psi_{t}}\sum_{i=1}^{N}\Context i{\tau}\Context i{\tau}^{T}\right)U_{t}^{-1}}\\
\le & \Trace{\left(\sum_{\tau\in\Psi_{t}}\sum_{i=1}^{N}\Context i{\tau}\Context i{\tau}^{T}+\lambda_{t}I\right)U_{t}^{-1}}\\
= & \Trace{I_{d}}=d.
\end{align*}
Thus, we have 
\[
\CP{\norm M_{F}>\frac{4N}{1-p}\sqrt{2d\log\frac{4t^{2}}{\delta}}}{\Psi_{t}}\le\frac{\delta}{t^{2}},
\]
and 
\begin{equation}
\begin{split}\Probability & \left(\norm{\sum_{\tau\in\Psi_{t}}\sum_{i=1}^{N}\left(1-\frac{\Indicator{\Tildeaction{\tau}=i}}{\SelectionP i{\tau}}\right)V_{t}^{-\frac{1}{2}}\XX i{\tau}\left(\Impute t-\beta^{*}\right)}_{2}>\frac{4}{1-p}\sqrt{2d\log\frac{4t^{2}}{\delta}}\right)\\
\le & \Probability\left(\frac{1}{N}\norm M_{F}>\frac{4}{1-p}\sqrt{2d\log\frac{4t^{2}}{\delta}}\right)\\
\le & \Expectation\CP{\norm M_{F}>\frac{4N}{1-p}\sqrt{2d\log\frac{4t^{2}}{\delta}}}{\Psi_{t}}\\
\le & \frac{\delta}{t^{2}}.
\end{split}
\label{eq:estimator_2nd_term}
\end{equation}

{[}Step 3. Bounding the second term in \eqref{eq:estimator_norm_decomposition}{]}
For the second term in \ref{eq:estimator_norm_decomposition}, we
have for any $x>0$, 
\begin{align*}
\Probability & \left(\norm{\sum_{\tau\in\Psi_{t}}\frac{\Error{\Tildeaction{\tau}}{\tau}}{\SelectionP{\Tildeaction{\tau}}{\tau}}\Context{\Tildeaction{\tau}}{\tau}+\sum_{\tau\notin\Psi_{t}}\Error{\Action{\tau}}{\tau}\Context{\Action{\tau}}{\tau}}_{V_{t}^{-1}}>x\right)\\
\le & \Probability\left(\left\{ \norm{\sum_{\tau\in\Psi_{t}}\frac{\Error{\Tildeaction{\tau}}{\tau}}{\SelectionP{\Tildeaction{\tau}}{\tau}}\Context{\Tildeaction{\tau}}{\tau}+\sum_{\tau\notin\Psi_{t}}\Error{\Action{\tau}}{\tau}\Context{\Action{\tau}}{\tau}}_{V_{t}^{-1}}>x\right\} \bigcap\left\{ \bigcap_{\tau\in\Psi_{t}}\left\{ \Tildeaction{\tau}=\Action{\tau}\right\} \right\} \right)\\
 & +\Probability\left(\bigcup_{\tau\in\Psi_{t}}\left\{ \Tildeaction{\tau}\neq\Action{\tau}\right\} \right)\\
\le & \Probability\left(\norm{\sum_{\tau\in\Psi_{t}}\frac{\Error{\Action{\tau}}{\tau}}{\SelectionP{\Action{\tau}}{\tau}}\Context{\Action{\tau}}{\tau}+\sum_{\tau\notin\Psi_{t}}\Error{\Action{\tau}}{\tau}\Context{\Action{\tau}}{\tau}}_{V_{t}^{-1}}>x\right).
\end{align*}
Since $\SelectionP{\Action{\tau}}{\tau}=p$, we observe that $\frac{\Error{\Action{\tau}}{\tau}}{\SelectionP{\Action{\tau}}{\tau}}$
and $\Error{\Action{\tau}}{\tau}$ are $\frac{\sigma}{p}$-sub-Gaussian.
Define $W_{t}:=\sum_{\tau=1}^{t}\Context{\Action{\tau}}{\tau}\Context{\Action{\tau}}{\tau}^{T}+\lambda_{t}I$.
Since $V_{t}\succeq W_{t}$, we have 
\[
\norm{\sum_{\tau\in\Psi_{t}}\frac{\Error{\Action{\tau}}{\tau}}{\SelectionP{\Action{\tau}}{\tau}}\Context{\Action{\tau}}{\tau}+\sum_{\tau\notin\Psi_{t}}\Error{\Action{\tau}}{\tau}\Context{\Action{\tau}}{\tau}}_{V_{t}^{-1}}\le\norm{\sum_{\tau\in\Psi_{t}}\frac{\Error{\Action{\tau}}{\tau}}{\SelectionP{\Action{\tau}}{\tau}}\Context{\Action{\tau}}{\tau}+\sum_{\tau\notin\Psi_{t}}\Error{\Action{\tau}}{\tau}\Context{\Action{\tau}}{\tau}}_{W_{t}^{-1}}.
\]
By assumption 2, $\Error{\Action{\tau}}{\tau}$ is a $\History{\tau+1}$-measurable
and $\sigma$-sub-Gaussian random variable given $\History{\tau}$.
Since $\Context{\Action{\tau}}{\tau}$ is $\History{\tau}$-measurable,
we can use Theorem 1 in \citet{abbasi2011improved} to have 
\begin{equation}
\norm{\sum_{\tau\in\Psi_{t}}\frac{\Error{\Action{\tau}}{\tau}}{p}\Context{\Action t}{\tau}+\sum_{\tau\notin\Psi_{t}}\Error{\Action{\tau}}{\tau}\Context{\Action{\tau}}{\tau}}_{W_{t}^{-1}}^{2}\le\frac{\sigma^{2}}{p^{2}}d\log\left(\frac{t}{\delta}\right),\label{eq:estimator_3rd_term}
\end{equation}
for all $t\ge0$ with probability at least $1-\delta$. Now with \eqref{eq:estimator_decompose}-\eqref{eq:estimator_3rd_term},
we can conclude that 
\begin{align*}
\norm{\Estimator t-\beta^{*}}_{V_{t}}\le & \frac{4}{1-p}\sqrt{2d\log\frac{4t^{2}}{\delta}}+\frac{\sigma}{p}\sqrt{d\log\left(\frac{t}{\delta}\right)}+\sqrt{\lambda_{t}}\\
\le & \left(\frac{4\sqrt{2}}{1-p}+\frac{\sigma}{p}\right)\sqrt{d\log\frac{4t^{2}}{\delta}}+\sqrt{\lambda_{t}},
\end{align*}
with probability at least $1-6\delta$. 
\end{proof}

\subsection{Proof of Lemma \ref{lem:psi_size}}

\label{sec:proof_of_phi_size} 
\begin{proof}
The proof follows from Chernoff's lower bound. In Algorithm \ref{alg:HyRan},
$\Psi_{t}$ is constructed as $\Psi_{t}=\{\tau\in[t]:\Tildeaction{\tau}=\Action{\tau}\}$.
Thus we have 
\[
\abs{\Psi_{t}}=\sum_{\tau=1}^{t}\Indicator{\Tildeaction{\tau}=\Action{\tau}}.
\]
Then for any $\epsilon\in(0,1)$ and $s<0$, 
\[
\Probability\left(\abs{\Psi_{t}}\le\epsilon pt\right)=\Probability\left(s\sum_{\tau=1}^{t}\Indicator{\Tildeaction{\tau}=\Action{\tau}}\ge s\epsilon pt\right)\le\exp\left(-s\epsilon pt\right)\Expectation\left[\exp\left(s\sum_{\tau=1}^{t}\Indicator{\Tildeaction{\tau}=\Action{\tau}}\right)\right].
\]
Let $\mathcal{G}_{\tau}=\Filtration{\tau}\cup\left\{ \Tildeaction 1,\ldots,\Tildeaction{\tau-1}\right\} $.
Then $\CE{\Indicator{\Tildeaction{\tau}=\Action{\tau}}}{\mathcal{G}_{\tau}}=p$,
for all $\tau\in[t]$ and 
\[
\begin{split}\Expectation\left[\exp\left(s\sum_{\tau=1}^{t}\Indicator{\Tildeaction{\tau}=\Action{\tau}}\right)\right] & =\Expectation\CE{\exp\left(s\sum_{\tau=1}^{t}\Indicator{\Tildeaction{\tau}=\Action{\tau}}\right)}{\mathcal{G}_{t}}\\
 & =\Expectation\left[\exp\left(s\sum_{\tau=1}^{t-1}\Indicator{\Tildeaction{\tau}=\Action{\tau}}\right)\CE{\exp\left\{ s\Indicator{\Tildeaction t=\Action t}\right\} }{\mathcal{G}_{t}}\right]\\
 & =\left\{ \left(1-p\right)+pe^{s}\right\} \Expectation\left[\exp\left(s\sum_{\tau=1}^{t-1}\Indicator{\Tildeaction{\tau}=\Action{\tau}}\right)\right]\\
 & =\vdots\\
 & =\left\{ \left(1-p\right)+pe^{s}\right\} ^{t}\\
 & \le\left\{ \exp\left(-p+pe^{s}\right)\right\} ^{t}.
\end{split}
\]
The last inequality holds due to $1+x\le e^{x}$ for all $x\in\Real$.
Thus, we have 
\[
\Probability\left(\abs{\Psi_{t}}\le\epsilon pt\right)\le\exp\left\{ \left(e^{s}-s\epsilon-1\right)pt\right\} .
\]
The right hand side is minimized when $s=\log\epsilon$. Setting $s=\log\epsilon$
gives 
\begin{align*}
\Probability\left(\abs{\Psi_{t}}\le\epsilon pt\right) & \le\exp\left\{ \left(\epsilon-\epsilon\log\epsilon-1\right)pt\right\} \le\exp\left\{ \left(-\epsilon^{2}+2\epsilon-1+\frac{(1-\epsilon)^{2}}{2}\right)pt\right\} ,\\
 & =\exp\left[\left\{ -\frac{1}{2}\left(1-\epsilon\right)^{2}\right\} pt\right]
\end{align*}
where the last inequality holds due to $\log x\ge x-1-\frac{(1-x)^{2}}{2x}$
for all $x\in(0,1)$. Setting the right hand side smaller than $\delta/T$
gives 
\begin{equation}
t\ge\frac{2}{p\left(1-\epsilon\right)^{2}}\log\frac{T}{\delta}.\label{eq:Psi_size_t}
\end{equation}
For $t$ that satisfies \eqref{eq:Psi_size_t}, $\Probability\left(\abs{\Psi_{t}}\le\epsilon pt\right)\le\frac{\delta}{T}$
holds. 
\end{proof}

\subsection{Proof of Theorem \ref{thm:lower_bound}}
\begin{proof}
The proof is inspired by that of Theorem 5.1 in \citet{auer2002nonstochastic},
and that of Theorem 24.2 in \citet{lattimore2020bandit}. Define the
context distribution $\mathcal{P}_{X}$ sampled from 
\[
\left(\begin{bmatrix}1\\
0\\
\vdots\\
0\\
0\\
\vdots\\
0
\end{bmatrix},\ldots,\begin{bmatrix}0\\
0\\
\vdots\\
0\\
0\\
\vdots\\
1
\end{bmatrix},\begin{bmatrix}0\\
0\\
\vdots\\
0\\
0\\
\vdots\\
0
\end{bmatrix}\right)\in\left(\Real^{d}\right)^{N}.
\]
Here, the covariance matrix $\Expectation\left[N^{-1}\sum_{i=1}^{N}\Context it\Context it\right]$
is positive definite. Let $\Error it$ be a random variable sampled
from the normal distribution $\mathcal{N}(0,1^{2})$, independently.
Then the reward distribution is Gaussian with mean $\Context it^{T}\beta$,
and variance $1^{2}$. For each $i\in[d]$ let $\beta_{i}=(0,\ldots0,\Delta,0\ldots,0)$
where $\Delta>0$ is in i-th component only. Then we have 
\begin{equation}
\Expectation_{\beta_{i}}\left[\sum_{t=1}^{T}\Context{\Optimalarm t}t^{T}\beta\right]=\Delta T.\label{eq:optimal_lower_bound}
\end{equation}
For each $i\in[d]$, we have 
\[
\Expectation_{\beta_{i}}\left[\sum_{t=1}^{T}\Context{\Action t}t^{T}\beta_{i}\right]=\Delta\Expectation_{\beta_{i}}\left[\sum_{t=1}^{T}\Indicator{\Action t=i}\right].
\]
Now set $\beta_{0}=\boldsymbol{0}$. Let $\Probability_{\beta_{i}}$
and $\Probability_{\beta_{0}}$ be the laws of $\sum_{t=1}^{T}\Indicator{\Action t=i}$
with respect to the bandit/learner interaction measure induced by
$\beta_{i}$ and $\beta_{0}$ respectively. Then by the result in
Exercise 14.4 in \citet{lattimore2020bandit}, 
\[
\Expectation_{\beta_{i}}\left[\sum_{t=1}^{T}\Indicator{\Action t=i}\right]\le\Expectation_{\beta_{0}}\left[\sum_{t=1}^{T}\Indicator{\Action t=i}\right]+T\sqrt{\frac{1}{2}D\left(\mathbb{P}_{\beta_{0}},\mathbb{P}_{\beta_{i}}\right)},
\]
where $D(\cdot,\cdot)$ is the relative entropy between two probability
measures. Set $\SetofContexts t:=(\Context 1t,\ldots,\Context Nt)$.
By the chain rule for the relative entropy, 
\[
\begin{split}D & \left(\mathbb{P}_{\beta_{0}},\mathbb{P}_{\beta_{i}}\right)\\
= & \sum_{t=1}^{T}D\left(\Probability_{\beta_{0}}\left(\Reward{\Action t}|\Reward{\Action 1},\ldots,\Reward{\Action{t-1}},\SetofContexts 1,\ldots,\SetofContexts t\right),\Probability_{\beta_{i}}\left(\Reward{\Action t}|\Reward{\Action 1},\ldots,\Reward{\Action{t-1}},\SetofContexts 1,\ldots,\SetofContexts t\right)\right)\\
 & +\sum_{t=1}^{T}D\left(\Probability_{\beta_{0}}\left(\SetofContexts t|\Reward{\Action 1},\ldots,\Reward{\Action{t-1}},\SetofContexts 1,\ldots,\SetofContexts{t-1}\right),\Probability_{\beta_{i}}\left(\SetofContexts t|\Reward{\Action 1},\ldots,\Reward{\Action{t-1}},\SetofContexts 1,\ldots,\SetofContexts{t-1}\right)\right)\\
= & \sum_{t=1}^{T}\Expectation_{\beta_{0}}\frac{\left\{ \Context{\Action t}t^{T}\left(\beta_{i}-\beta_{0}\right)\right\} ^{2}}{2}\\
= & \frac{\Delta^{2}}{2}\Expectation_{\beta_{0}}\left[\sum_{t=1}^{T}\Indicator{\Action t=i}\right],
\end{split}
\]
where the second equality holds since the distribution of $\SetofContexts t$
does not change over $\beta$, and 
\[
\begin{split}D & \left(\Probability_{\beta_{0}}\left(\Reward{\Action t}|\Reward{\Action 1},\ldots,\Reward{\Action{t-1}},\SetofContexts 1,\ldots,\SetofContexts t\right),\Probability_{\beta_{i}}\left(\Reward{\Action t}|\Reward{\Action 1},\ldots,\Reward{\Action{t-1}},\SetofContexts 1,\ldots,\SetofContexts t\right)\right)\\
 & =\int\int\log\frac{d\Probability_{\beta_{i}}(y|a_{t})}{d\Probability_{\beta_{0}}(y|a_{t})}d\Probability_{\beta_{0}}(y|a_{t})d\Probability_{\beta_{0}}\left(a_{t}\right)\\
 & =\int\frac{\left\{ \Context{\Action t}t^{T}\left(\beta_{i}-\beta_{0}\right)\right\} ^{2}}{2}d\Probability_{\beta_{0}}\left(a_{t}\right)\\
 & =\Expectation_{\beta_{0}}\frac{\left\{ \Context{\Action t}t^{T}\left(\beta_{i}-\beta_{0}\right)\right\} ^{2}}{2}.
\end{split}
\]
Thus we have 
\[
\Expectation_{\beta_{i}}\left[\sum_{t=1}^{T}\Context{\Action t}t^{T}\beta_{i}\right]\le\Delta\Expectation_{\beta_{0}}\left[\sum_{t=1}^{T}\Indicator{\Action t=i}\right]+\frac{\Delta^{2}T}{2}\sqrt{\Expectation_{\beta_{0}}\left[\sum_{t=1}^{T}\Indicator{\Action t=i}\right]}.
\]
With \eqref{eq:optimal_lower_bound}, 
\[
\Expectation_{\beta_{i}}\left[R(T)\right]\ge\Delta T-\Delta\Expectation_{\beta_{0}}\left[\sum_{t=1}^{T}\Indicator{\Action t=i}\right]-\frac{\Delta^{2}T}{2}\sqrt{\Expectation_{\beta_{0}}\left[\sum_{t=1}^{T}\Indicator{\Action t=i}\right]}.
\]
Taking average over $i\in[d]$ gives 
\[
\begin{split}\frac{1}{d}\sum_{i=1}^{d}\Expectation_{\beta_{i}}\left[R(T)\right] & \ge\Delta T-\frac{\Delta}{d}\sum_{i=1}^{d}\Expectation_{\beta_{0}}\left[\sum_{t=1}^{T}\Indicator{\Action t=i}\right]-\frac{\Delta^{2}T}{2d}\sum_{i=1}^{d}\sqrt{\Expectation_{\beta_{0}}\left[\sum_{t=1}^{T}\Indicator{\Action t=i}\right]}\\
 & \ge\Delta T-\frac{\Delta T}{d}-\frac{\Delta^{2}T\sqrt{d}}{2d}\sqrt{\sum_{i=1}^{d}\Expectation_{\beta_{0}}\left[\sum_{t=1}^{T}\Indicator{\Action t=i}\right]}\\
 & \ge\frac{\Delta T}{2}-\frac{\Delta^{2}T\sqrt{T}}{2\sqrt{d}}.
\end{split}
\]

Setting $\Delta=\frac{1}{2}\sqrt{\frac{d}{T}}$ gives 
\[
\frac{1}{d}\sum_{i=1}^{d}\Expectation_{\beta_{i}}\left[R(T)\right]\ge\frac{1}{8}\sqrt{dT}.
\]
Thus, there exists $\beta_{i}$ such that $\Expectation_{\beta_{i}}\left[R(T)\right]\ge\frac{1}{8}\sqrt{dT}$. 
\end{proof}

\section{LIMITATIONS}

\label{sec:limitations}
\begin{enumerate}
\item The regret bound is derived under stochastic conditions for contexts
in Assumption~\ref{assum:iid_context}. Although the same or similar
assumptions have been used in the previous literature~\citep{li2017provably,amani2019linear,oh2021sparsity,kim2021doubly},
we hope that this can be relaxed in the future work. Nevertheless,
achieving a regret bound sublinear in both time horizon and the dimensionality,
even under such a stochastic assumption, has not been shown for any
practical algorithm other than the variants of ``\texttt{Sup}''-type
algorithms~\citep{auer2002using}. We strongly believe that our work
fills the long-standing gap between sublinear dependence on $d$ and
a practical algorithm other than \texttt{SupLinUCB} variants. 
 
\item The proposed Hyran estimator requires more computations compared to
ridge estimator in that it uses contexts of all arms and the imputation
estimator. However, we believe that these additional computations
are reasonable costs to obtain more precise estimator and to achieve
a near-optimal regret bound. 
\end{enumerate}

\end{document}